\documentclass[10pt,twocolumn,letterpaper]{article}

\usepackage{cvpr}              
\usepackage{booktabs}
\usepackage{subcaption}
\usepackage{array}
\usepackage{longtable}
\usepackage{multirow}
\usepackage{tabularx}

\newcommand{\y}{\textbf{Y}}
\newcommand{\n}{\textbf{N}}
\newcommand{\p}{\textbf{P}}
\usepackage{pifont}           
\usepackage{amssymb}          
\usepackage{algorithm}        
\usepackage{algpseudocode}
\usepackage{xcolor}
\usepackage{setspace}
\newcommand{\cmark}{\ding{51}}
\newcommand{\xmark}{\ding{55}}

\definecolor{darkgreen}{rgb}{0.0, 0.5, 0.0}
\definecolor{darkred}{rgb}{0.8, 0.0, 0.0}
\newcommand{\yes}{\textcolor{darkgreen}{\cmark}}
\newcommand{\no}{\textcolor{darkred}{\xmark}}

\definecolor{cvprblue}{rgb}{0.21,0.49,0.74}
\usepackage[pagebackref,breaklinks,colorlinks,allcolors=cvprblue]{hyperref} 

\def\paperID{4079} 
\def\confName{CVPR}
\def\confYear{2026}

\title{TESSERA: Temporal Embeddings of Surface Spectra for\\ Earth Representation and Analysis}

\author{
Zhengpeng Feng$^1$ \quad Clement Atzberger$^{2\dagger}$ \quad Sadiq Jaffer$^{1\dagger}$ \quad Jovana Knezevic$^1$ \quad Silja Sormunen$^3$ \\
Robin Young$^1$ \quad Madeline C. Lisaius$^1$ \quad Markus Immitzer$^2$ \quad Toby Jackson$^4$ \quad James Ball$^1$ \\
David A. Coomes$^1$ \quad Anil Madhavapeddy$^1$ \quad Andrew Blake$^1$ \quad Srinivasan Keshav$^{1*}$ \\
$^1$University of Cambridge \quad $^2$dClimate Labs \quad  $^3$Aalto University \quad $^4$University of Bristol \\
{\tt\small$^\dagger$ Equal contribution.
\small \{zf281, sj514, ray25, mcl66, avsm2, sk818\}@cam.ac.uk}
}

\begin{document}
\maketitle
\def\thefootnote{*}\footnotetext{Corresponding author.}

\allowdisplaybreaks

\begin{abstract}
Satellite Earth-observation (EO) time series in the optical and microwave ranges of the electromagnetic spectrum are often irregular due to orbital patterns and cloud obstruction. Compositing addresses these issues but loses information with respect to vegetation phenology, which is critical for many downstream tasks. Instead, we present TESSERA, a pixel-wise foundation model for multi-modal (Sentinel-1/2) EO time series that learns robust, label-efficient embeddings. During model training, TESSERA uses Barlow Twins and sparse random temporal sampling to enforce invariance to the selection of valid observations. We employ two key regularizers: global shuffling to decorrelate spatial neighborhoods and mix-based regulation to improve invariance under extreme sparsity.
We find that for diverse classification, segmentation, and regression tasks, TESSERA embeddings deliver state-of-the-art accuracy with high label efficiency, often requiring only a small task head and minimal computation.  To democratize access, adhere to FAIR principles, and simplify use, we
release global, annual, 10m, pixel-wise int8 embeddings together with open weights/code and lightweight adaptation heads, thus providing practical tooling for large-scale retrieval and inference at planetary scale. 
All code and data are available at \url{https://github.com/ucam-eo/tessera}.

\end{abstract}

\section{Introduction}
\label{sec:intro}
\begin{figure*}
    \centering
    \includegraphics[width=1\linewidth]{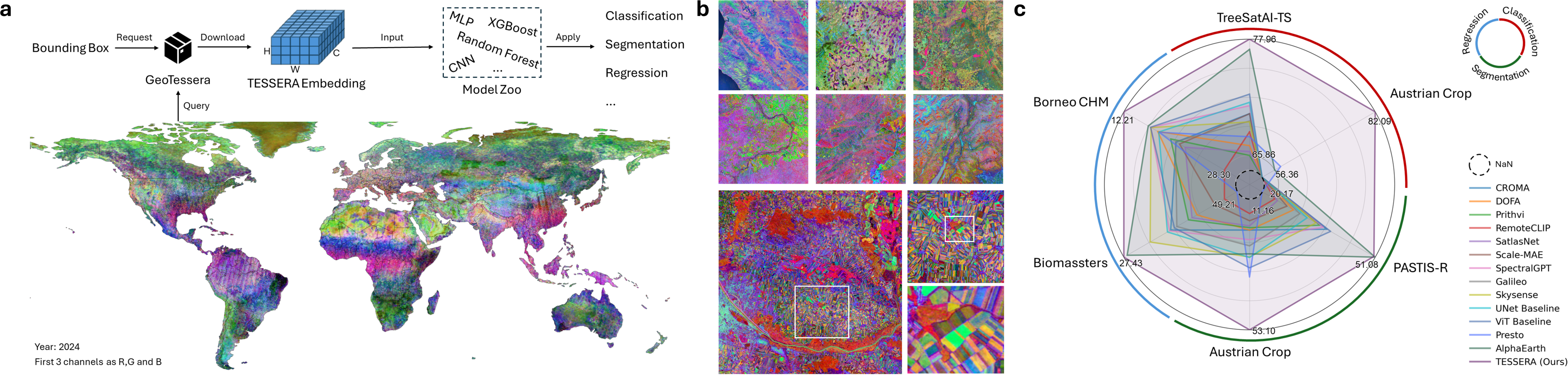}
    \caption{\textbf{TESSERA advances the embedding-as-data approach for Earth Observation.}
    It delivers \textbf{(a)} analysis-ready global-scale products, \textbf{(b)} pixel-wise, high-fidelity representations, and has \textbf{(c)} state-of-the-art downstream accuracy.}
    \label{fig:main_figure}
    \vspace{-3mm}
\end{figure*}

Earth Observation (EO) satellites continuously observe our planet, yet cloud cover, irregular revisit intervals depending on latitude, sensors that differ in repeat period, spatial and spectral resolutions, and noise complicate the resulting time series. For many EO applications, such as agriculture, forestry, and environmental monitoring, clouds frequently obscure the evolution of target phenomena. Although compositing can yield cloud-free mosaics and reduce noise in microwave measurements, it flattens phenological dynamics and transient events that downstream models require~\cite{zeng2020review}.

Recent remote-sensing foundation models (RSFMs) have made substantial progress in large-scale pretraining, yet most assume heavily preprocessed cloud-filtered inputs. These predominantly patch-based models train on composites or temporal averages, a simplification that discards partly-clouded observations that contain valuable temporal information. Consequently, learned embeddings often bias toward idealized, cloud-free conditions, limiting their robustness to the irregular, incomplete sampling that characterizes operational EO time series~\cite{xiao2025foundation}.

We approach learning EO-based embeddings differently: rather than filtering imperfect observations, we use Barlow Twins~\cite{Zbontar2021BarlowTS, Lisaius2024UsingBT} to create embeddings that are robust to temporal sampling variability. Specifically, by forcing embeddings to be invariant to randomly selected subsets of cloud-free observations from the same location, our embeddings model consistent physical processes, which results in their generalization across sensors, seasons, and regions.

Building on work by Lisaius \textit{et al.}~\cite{Lisaius2024UsingBT}, which describes the
use of Barlow Twins for EO for optical data streams, 
we introduce TESSERA, a pixel-wise RSFM for multi-modal (Sentinel-1/2) sequences.
Our work differs from that of Lisaius \textit{et al.}~\cite{Lisaius2024UsingBT} in four additional ways. 
First,
their work created embeddings from Sentinel 2 alone,  
whereas TESSERA combines observations from Sentinel 1 and 2 in
a novel architecture.
Second, we introduce two complementary regularizers: global shuffling to disrupt short-range spatial correlations, while mix-based regularization blends observations across time and samples to stabilize learning under severe sparsity. 

Third, unlike nearly all prior work,
open-source global, annual, pixel-wise int8 embeddings for each 10m $\times$ 10m location on land, along with open-source model weights, code, and the \texttt{GeoTessera} Python library~\cite{ucam-eo-geotessera} to efficiently retrieve embeddings from our data store for large-scale inference. 
TESSERA embeddings can be paired with lightweight adaptation heads (small MLPs or shallow convolutional models) to achieve high performance with minimal computation, enabling rapid prototyping and straightforward deployment. TESSERA thus emphasizes usability through the FAIR principles (Findable, Accessible, Interoperable, and Reusable). 

Finally, 
we evaluate TESSERA on a wide range of classification, segmentation, and regression benchmarks spanning multiple regions and label regimes. 
We find that our embeddings remain consistent when calendars shift, seasons are partially missing, or cloud cover is dense.
Moreover, we achieve state-of-the-art accuracy with strong label efficiency, particularly under low-label conditions, and the TESSERA embeddings degrade gracefully under heavy cloud cover. Ablations confirm that temporal sampling invariance, global shuffling, and mix-based regulation contribute each to robustness and generalization.

Our contributions are:
\begin{enumerate}
    \item \textbf{Global, pixel-wise, label-efficient embeddings.} We use a novel self-supervised architecture to train a multi-modal (S1/S2) pixel-wise RSFM.
    \item \textbf{FAIR embeddings-as-data.} We are releasing global, annual, 10\,m, pixel-wise int8 embeddings with open-source weights, code, and a Python library—advancing FAIR deployment-ready resources.
     \item \textbf{Comprehensive evaluation.} We compare our embeddings with a large number of other state-of-the-art foundation models and find that, for diverse classification, segmentation, and regression tasks, TESSERA embeddings deliver outstanding performance.
\end{enumerate}

\section{Related Work}
\label{sec:related_work}

\subsection{Foundation Models for Earth Observation}
Remote-sensing foundation models (RSFMs) have rapidly advanced along several methodological lines. A large body of work has used self-supervised pretraining using either contrastive objectives (e.g., SeCo~\cite{Manas2021}, SkySense~\cite{Guo2024}) or masked image modeling (e.g., SatMAE~\cite{Cong2022}, Scale-MAE~\cite{Reed2023}, Cross-Scale MAE~\cite{Tang2023}, S2MAE~\cite{li2024s2mae}, SatMAE++~\cite{Noman2024}). Other variants leverage global–local view alignment or distilled objectives (e.g., DINO-MM/DINO-MC~\cite{wang2022selfsupervisedvisiontransformersjoint,wanyan2024extendinggloballocalviewalignment}). Model scopes have expanded from single-modality encoders toward broader families such as RingMo and RingMo-Lite~\cite{Sun2023,wang2023ringmoliteremotesensingmultitask}, Satlas~\cite{Bastani2023}, Prithvi and Prithvi-EO-2.0~\cite{schmude2024prithviwxcfoundationmodel,szwarcman2025prithvieo20versatilemultitemporalfoundation}, GFM/msGFM~\cite{Mendieta2023,Han2024}, Skysense series \cite{Guo2024, wu2025semantic,luo2024skysensegpt,zhu2025skysense,zhang2025skysense}. Work on multi-sensor fusion explicitly integrates optical and SAR (e.g., CROMA~\cite{fuller2023croma}, SkySense~\cite{Guo2024}, RingMo-Sense~\cite{10254320}). In parallel, early efforts like MOSAIKS~\cite{rolf2021generalizable} explored lightweight universal features to lower adoption barriers.

Despite these advances, most RSFMs are trained on patch-based input and implicitly assume that the data has been preprocessed (e.g., aggressive cloud filtering or temporal averaging). Such assumptions simplify pretraining, but discard partially cloudy images that still carry useful temporal signals, biasing representations toward idealized, cloud-free conditions. Moreover, strong performance typically requires fine-tuning the backbone for each downstream task, which can be compute- and label-intensive for EO users.

\subsection{Embedding-Based Foundation Models}
\begin{figure}[t]
    \centering
    \includegraphics[width=1\linewidth]{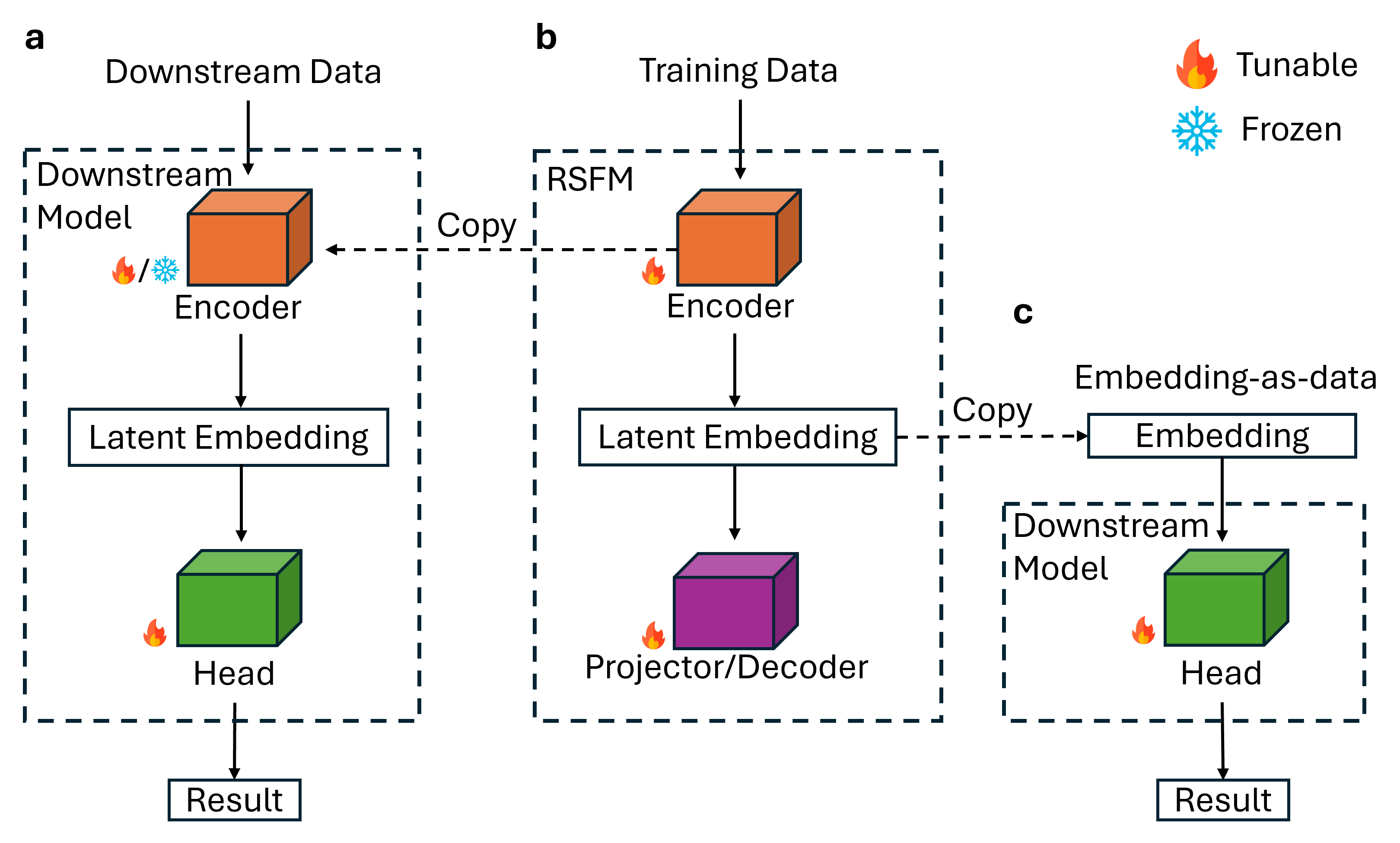}
    \caption{\textbf{Downstream adaptation paradigms.}
    \textbf{(a–b)} Most RSFMs require fine-tuning the encoder or training task-specific decoders and often do not expose intermediate embeddings.
    \textbf{(b-c)} \textbf{TESSERA} produces fixed, task-agnostic embeddings that plug into lightweight heads, avoiding backbone fine-tuning.}
    \label{fig:downstream_paradigm}
    \vspace{-3mm}
\end{figure}

A complementary line of work delivers \emph{embeddings-as-data}: precomputed, generic, task-agnostic features that can be easily adapted with lightweight heads (shown in~\Cref{fig:downstream_paradigm} c). Representative models include Presto~\cite{tseng_lightweight_2024}, which learns pixel-level temporal representations, and Google's AlphaEarth~\cite{SatelliteEmbeddingV1}, which provides precomputed embeddings as a ready-for-analysis product. This paradigm reduces computational costs and democratizes access to high performance; however, existing approaches are often not fully open-sourced, lack pixel-level granularity at a global scale, or do not explicitly account for irregular temporal sampling.

Our work follows the embedding-based paradigm while addressing these gaps. TESSERA introduces temporal sampling invariance to handle cloud- and orbital pattern induced irregularity and combines redundancy reduction with global shuffling and mix-based regulation for robustness. In contrast to prior embedding releases, we provide FAIR global, annual, 10\,m pixel-wise embeddings, together with open weights, code, and the \texttt{GeoTessera} library~\cite{ucam-eo-geotessera} enabling analysis-ready use without backbone fine-tuning.

\section{TESSERA}
\label{sec:method}

\begin{figure*}[t]
    \centering
    \includegraphics[width=\linewidth]{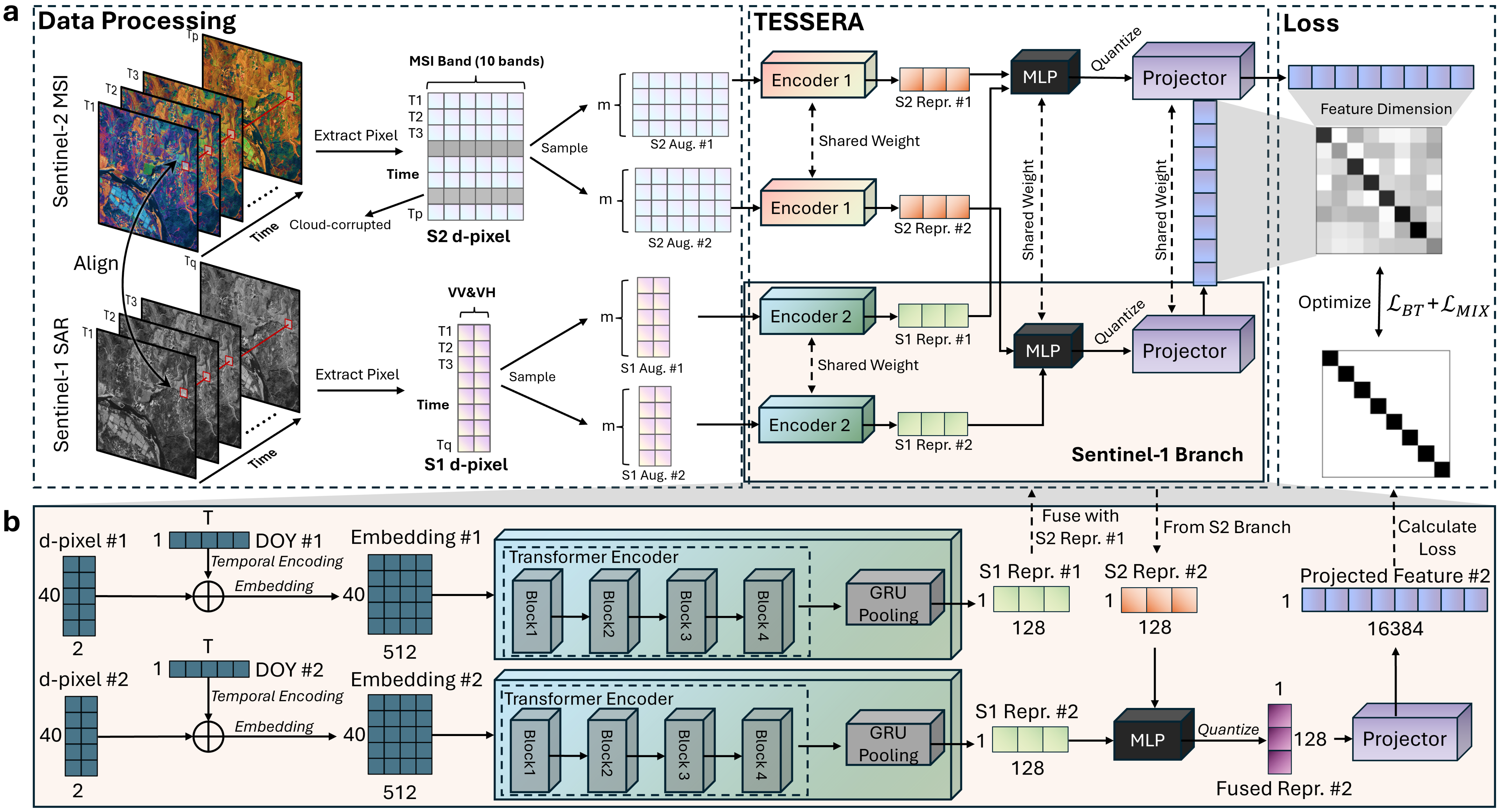}
    \caption{\textbf{Overview of the TESSERA processing pipeline.}
    \textbf{(a)} Overall architecture: multi-temporal Sentinel-1/2 observations are converted into modality-specific \emph{d-pixels}, augmented twice, and encoded by dual branches with shared weights to produce compact 128-D embeddings.
    \textbf{(b)} Zoom-in on the Sentinel-1 branch: temporal sequences are processed by a 4-block Transformer followed by GRU pooling to capture dynamic backscatter patterns before fusion into the joint TESSERA embedding.}
    \label{fig:model_architecture}
    \vspace{-3mm}
\end{figure*}

\subsection{d-pixel: A Representation for Temporal Data}

We consider remote sensing data~\cite{Sentinel1_Data, Sentinel2_Data} with $C$ channels (spectral bands or polarizations). Each data tile $R_t$ at time $t$ is represented as a 3D array with dimensions:
{
\setlength{\abovedisplayskip}{3pt}
\setlength{\belowdisplayskip}{3pt}
\[
R_t \in \mathbb{R}^{W \times H \times C}
\]
}
where $W$ is the width (longitude dimension), $H$ is the height (latitude dimension), and $C$ is the number of spectral channels.

Each tile is accompanied by a corresponding binary mask $V_t$ of dimensions:
{
\setlength{\abovedisplayskip}{3pt}
\setlength{\belowdisplayskip}{3pt}
\[
V_t \in \{0,1\}^{W \times H}
\]
}
where $V_t(i,j) = 0$ indicates clouding or missing data for the pixel at spatial coordinates $(i,j)$, and $V_t(i,j) = 1$ indicates valid data.

\vspace{-3mm}
\paragraph{Temporal Data Stacking}
We stack resampled and spatially-aligned tiles over a time period that spans $T$ time steps ($t = 0, 1, \ldots, T-1$). The temporal data stack is defined as
{
\setlength{\abovedisplayskip}{3pt}
\setlength{\belowdisplayskip}{3pt}
\begin{align*} 
\mathbf{D} = [R_0, R_1, \ldots, R_{T-1}] \\
\mathbf{M} = [V_0, V_1, \ldots, V_{T-1}]
\end{align*}
}

\paragraph{Time Series Extraction and d-pixel Definition}
For a given spatial location $(i,j)$ and spectral channel $c$, the time series $S_{i,j,c}$ represents all channel $c$ values at coordinates $(i,j)$ over the entire time period:
{
\setlength{\abovedisplayskip}{3pt}
\setlength{\belowdisplayskip}{3pt}
\[
S_{i,j,c} = [R_0(i,j,c), R_1(i,j,c), \ldots, R_{T-1}(i,j,c)]
\]
}
We define a \textit{d-pixel} $P_{i,j}$ as the collection of all spectral channels by timesteps at a given spatial location $(i,j)$:
{
\setlength{\abovedisplayskip}{3pt}
\setlength{\belowdisplayskip}{3pt}
\[
P_{i,j}(c) = S(i,j, c)
\]
}
In other words, the d-pixel provides all spectral values (Sentinel-2) or backscatter values (Sentinel-1) at a given point over time. Note that d-pixels are usually sparse and are accompanied by a mask vector $m_{i,j}$ of size $T$ that indicates the timesteps for which there are valid data, with a value of 1 indicating that the corresponding row in $P_{i,j}$ is valid.
The d-pixel formulation preserves the full temporal phenology while gracefully accommodating irregular sampling.

\subsection{Model Architecture}
TESSERA employs a simple yet efficient dual-branch encoder architecture to process optical and radar modalities separately before fusion.

\vspace{-3mm}
\paragraph{Modality Encoders.}
Each encoder receives a masked time series $\{ (P_{i,j}^{(t)}, m_{i,j}^{(t)}) \}_{t=1}^T$. Valid observations are embedded via a linear projection $\phi: \mathbb{R}^C \to \mathbb{R}^d$, and temporal context is injected via learnable Day-of-Year (DoY) positional encodings:
{
\setlength{\abovedisplayskip}{3pt}
\setlength{\belowdisplayskip}{3pt}
\[
e_t = \phi(P_{i,j}^{(t)}) + \psi(\text{DoY}(t)),
\]
}
where $\psi(\cdot)$ maps the normalized day of the year to a $d$-dimensional embedding space. The sequence $\{e_t\}_{t: m_{i,j}^{(t)}=1}$ is padded to a fixed length $L=40$ via sampling with replacement if necessary, then processed by a 4-layer Transformer encoder with multi-head self-attention:
{
\setlength{\abovedisplayskip}{3pt}
\setlength{\belowdisplayskip}{3pt}
\[
H = \text{TransformerEncoder}([e_{t_1}, \dots, e_{t_L}]).
\]
}
A GRU-based pooling layer aggregates the temporal sequence into a fixed-size vector:
{
\setlength{\abovedisplayskip}{3pt}
\setlength{\belowdisplayskip}{3pt}
\[
z_{\text{mod}} = \text{GRU}(H) \in \mathbb{R}^{128}.
\]
}

\vspace{-3mm}
\paragraph{Feature Fusion and Quantization.}
The optical embeddings ($z_{\text{S2}}$) and radar embeddings ($z_{\text{S1}}$) are concatenated and passed through a 2-layer MLP to produce the fused 128-dimensional embedding:
{
\setlength{\abovedisplayskip}{3pt}
\setlength{\belowdisplayskip}{3pt}
\[
z = \text{MLP}([z_{\text{S2}}; z_{\text{S1}}]) \in \mathbb{R}^{128}.
\]
}
We apply Quantization-Aware Training (QAT) to compress $z$ into 8-bit integers, reducing storage by $\sim$4$\times$ with negligible performance loss.

\vspace{-3mm}
\paragraph{Projector Network.}
For training, $z$ is expanded to 16,384 dimensions using a deep projector MLP:
{
\setlength{\abovedisplayskip}{3pt}
\setlength{\belowdisplayskip}{3pt}
\[
\hat{z} = \text{Projector}(z) \in \mathbb{R}^{16384},
\]
}
which is discarded during inference.

\subsection{Pretraining}
TESSERA is pretrained on $\sim$800 million d-pixels sampled from 3,012 global MGRS tiles (2017–2024), using 16 AMD MI300X GPUs (192GB GPU RAM each). For each d-pixel, two augmented views $(Y_A, Y_B)$ are generated by independently sampling 40 valid time steps from the Sentinel-1 and Sentinel-2 time series. The model is trained for one epoch with a global batch size of 32,768 using AdamW ($\eta=0.002$, weight decay $10^{-6}$), with a linear warmup followed by cosine decay.

The network processes these two views ($Y_A, Y_B$) through the dual-encoder and the projector to produce batch-normalized embeddings $Z_A$ and $Z_B$. The standard Barlow Twins loss function, $\mathcal{L}_{BT}$, is defined as~\cite{Zbontar2021BarlowTS}:
\begin{equation*} \label{eq:barlow_twins_loss_appendix}
\mathcal{L}_{BT} = \sum_i (1 - C_{ii})^2 + \lambda_{BT} \sum_i \sum_{j \neq i} C_{ij}^2
\end{equation*}

Here, $C$ is the cross-correlation matrix computed between the batch-normalized embeddings $Z_A$ and $Z_B$. The indices $i$ and $j$ iterate over the dimensions of the embedding vectors. The first term (invariance term) encourages similar embeddings for different views of the same input ($C_{ii} \to 1$). The second term (redundancy reduction term) promotes information efficiency by minimizing the correlation between different embedding dimensions ($C_{ij} \to 0$ for $i \neq j$), weighted by $\lambda_{BT}$. We used the value of $\lambda_{BT} = 5 \times 10^{-3}$ as recommended in Reference~\cite{Zbontar2021BarlowTS}.

To further enhance the robustness of the model and mitigate overfitting, TESSERA incorporates an additional mix-up regularization term $\mathcal{L}_{MIX}$, inspired by Bandara et al.~\cite{Bandara2023MixCoAM}. For each training batch, this involves shuffling one set of views (e.g., $Y_B$) along the batch dimension to create $Y_S = \text{Shuffle}(Y_B)$, then generating a mixed view $Y_M = \alpha_{mix}Y_A + (1-\alpha_{mix})Y_S$. The mixing coefficient $\alpha_{mix}$ is sampled from a uniform distribution in the unit interval $\alpha_{mix} \sim U(0, 1)$. The embeddings $Z_M$ and $Z_S$ are obtained from their respective views. The mix-up loss penalizes deviations from the assumption that a linear interpolation in the input space corresponds to a linear interpolation in the embedding space.
\begin{align*} 
C_{target}^{MA} &= \alpha_{mix}(Z_{A})^{T}Z_{A}+(1-\alpha_{mix})(Z_{S})^{T}Z_{A} \\
C_{target}^{MS} &= \alpha_{mix}(Z_{A})^{T}Z_{S}+(1-\alpha_{mix})(Z_{S})^{T}Z_{S} \\
\mathcal{L}_{MIX} &= \frac{1}{2}(\|C^{MA}-C_{target}^{MA}\|_{F}^{2}+\|C^{MS}-C_{target}^{MS}\|_{F}^{2})
\end{align*}
where $C^{MA}=(Z_{M})^{T}Z_{A}$ and $C^{MS}=(Z_{M})^{T}Z_{S}$ are the actual cross-correlation matrices from the model output. The total loss function optimized during training is a weighted sum:
\begin{equation*} \label{eq:total_loss}
\mathcal{L}_{total} = \mathcal{L}_{BT} + \lambda_{mix}\mathcal{L}_{MIX}
\end{equation*}
where $\lambda_{mix}$ controls the strength of the mix-up regularization. We found $\lambda_{mix} = 1.0$ (as recommended by~\cite{Bandara2023MixCoAM}) to be effective based on our experiments.

Crucially, we implement a \textit{global shuffling} strategy that randomizes d-pixels across all geographic tiles before batching, smoothing the loss curve, preventing spatial autocorrelation, and improving generalization (see the supplementary materials).

\subsection{Inference}
Following pretraining, the TESSERA dual encoder (with frozen weights and excluding the projector) is used to generate an annual 128-dimensional embedding for every 10m pixel globally. This involves making inferences from a sample of length 40 from the annual Sentinel-1 and Sentinel-2 time series for each pixel through the trained encoders. To ensure that our maps are gap-free, random sampling with replacement is used
to generate augmentations from d-pixels that have fewer than 40 non-cloudy observations. The primary output of this inference stage is a set of annual, global, 10m resolution TESSERA embedding maps. These maps are designed as readily usable multidimensional geospatial data layers. This ``Embeddings-as-Data'' approach significantly lowers the barrier to entry for end-users, as these rich, precomputed features can be directly ingested by downstream models without the need for raw satellite data processing, running the TESSERA model itself (see supplementary materials). 
We release embeddings through the pip-installable \texttt{GeoTessera} Python library, enabling easy access and integration into downstream workflows.
\section{Experiments}
\label{sec:experiments}

\begin{table*}[t]
\centering
\caption{\textbf{Results across tasks.}
\textbf{(a) Classification}: F1 $\uparrow$ on TreeSatAI-TS
and Austrian Crop (left column; TreeSatAI-TS shows
\textit{original/finetuned}).
\textbf{(b) Segmentation}: mIoU $\uparrow$ on PASTIS-R and
Austrian Crop (\textit{original/finetuned} for CROMA--ViT Baseline; single
evaluation for Presto, AlphaEarth, TESSERA).
\textbf{(c) Regression}: RMSE $\downarrow$ on Biomassters
and Borneo CHM (\textit{original/finetuned} for baselines; single evaluation
otherwise).
\textbf{best}; \underline{second}.}
\label{tab:unified_result_table}
\setlength{\tabcolsep}{5pt}
\renewcommand{\arraystretch}{1.05}

\begin{minipage}[t]{0.48\textwidth}
\centering
\scriptsize
\textbf{(a1) Classification --- TreeSatAI-TS (F1
$\uparrow$)}\\[2pt]
\resizebox{\linewidth}{!}{
\begin{tabular}{l|ccc}
\toprule
\multirow{2}{*}{\textbf{Model}} &
\multicolumn{3}{c}{\textbf{TreeSatAI-TS}} \\
& 1\% & 30\% & All \\
\midrule
CROMA\cite{fuller2023croma} & 39.73/43.14 & 62.44/63.24 & 70.21/72.09
\\
DOFA\cite{ xiong2024neuralplasticityinspiredmultimodalfoundation} & 35.04/40.15 & 59.23/63.46 & 66.88/67.20
\\
Prithvi\cite{schmude2024prithviwxcfoundationmodel} & 37.21/41.89 & 60.01/60.66 & 65.86/67.04
\\
RemoteCLIP\cite{remoteclip} & 34.94/37.77 & 57.52/59.70 & 68.32/68.03
\\
SatlasNet\cite{Bastani2023} & 33.86/37.83 & 56.80/61.31 & 66.27/67.85
\\
Scale-MAE\cite{Reed2023} & 36.20/40.96 & 59.26/61.29 & 70.20/73.12
\\
SpectralGPT\cite{SpectralGPT} & 41.82/45.66 & 59.77/60.64 & 71.04/72.23
\\
Galileo\cite{tseng2025galileolearningglobal} & 39.99/42.38 & 59.42/61.94 & 69.44/72.17
\\
Skysense\cite{Guo2024} & 39.72/41.39 & 61.88/64.80 & 69.35/72.24
\\
\midrule
UNet Baseline\cite{ronnebergerUNetConvolutionalNetworks2015} & 39.88/40.43 & 60.03/62.22 & 71.39/73.30
\\
ViT Baseline\cite{dosovitskiy2020image} & 40.52/41.70 & 61.79/65.62 & 72.24/72.60
\\
\midrule
Presto\cite{tseng_lightweight_2024} & 46.02 & 61.45 & 67.81 \\
AlphaEarth\cite{SatelliteEmbeddingV1} & \underline{52.79} & \underline{71.94} & \underline{76.90} \\
\textbf{TESSERA (Ours)} & \textbf{60.58} &
\textbf{75.42} & \textbf{77.96} \\
\bottomrule
\end{tabular}
}
\vspace{6pt}
\hrule
\vspace{6pt}
\textbf{(a2) Classification --- Austrian Crop (F1
$\uparrow$)}\\[2pt]
\resizebox{\linewidth}{!}{
\begin{tabular}{l|cccccccc}
\toprule
\multirow{2}{*}{\textbf{Model}} &
\multicolumn{8}{c}{\textbf{Austrian Crop}} \\
& 1\% & 3\% & 5\% & 7\% & 10\% &
15\% & 20\% & 30\% \\
\midrule
Presto\cite{tseng_lightweight_2024} & 32.74 & 37.63 & 40.26 & 41.50 & 43.94 & 46.01 & 49.74 & \underline{57.89} \\
AlphaEarth\cite{SatelliteEmbeddingV1} & \underline{37.22} & \underline{40.06} & \underline{42.73} & \underline{45.08} & \underline{48.01} & \underline{50.47} & \underline{52.47} & 56.36 \\
\textbf{TESSERA (Ours)} & \textbf{66.15} & \textbf{71.62} & \textbf{74.32} & \textbf{77.22} & \textbf{78.69} & \textbf{80.12} & \textbf{80.79} & \textbf{82.09} \\
\bottomrule
\end{tabular}
}
\end{minipage}
\hfill
\begin{minipage}[t]{0.5\textwidth}
\centering
\scriptsize
\textbf{(b) Segmentation --- PASTIS-R / Austrian Crop (mIoU
$\uparrow$)}\\[2pt]
\resizebox{\linewidth}{!}{
\begin{tabular}{l|ccc|ccc}
\toprule
\multirow{2}{*}{\textbf{Model}} &
\multicolumn{3}{c|}{\textbf{PASTIS-R}} & \multicolumn{3}{c}{\textbf{Austrian
Crop}} \\
& 1\% & 30\% & All & 1\% & 30\% &
All \\
\midrule
CROMA\cite{fuller2023croma} & 7.61/12.33 & 21.62/23.78 & 36.24/41.29
& 5.64/5.19 & 12.89/15.76 & 17.10/18.28 \\
DOFA\cite{xiong2024neuralplasticityinspiredmultimodalfoundation} & 4.82/5.53 & 19.28/21.74 & 25.94/29.55
& 4.90/4.92 & 10.05/14.00 & 17.14/17.39 \\
Prithvi\cite{schmude2024prithviwxcfoundationmodel} & 8.24/11.01 & 24.86/29.51 & 34.09/39.52
& 4.36/3.82 & 11.12/14.35 & 16.25/18.60 \\
RemoteCLIP\cite{remoteclip} & 5.50/7.55 & 11.72/16.40 & 20.17/21.03
& 3.52/3.37 & 7.76/10.12 & 11.16/15.15 \\
SatlasNet\cite{Bastani2023} & 9.54/11.58 & 17.86/21.48 & 21.02/31.38
& 4.27/4.05 & 8.45/9.64 & 15.61/18.97 \\
Scale-MAE\cite{Reed2023} & 5.57/7.86 & 16.34/18.48 & 23.96/28.51
& 4.48/4.93 & 7.42/10.31 & 13.75/15.22 \\
SpectralGPT\cite{SpectralGPT} & 11.25/15.95 & 26.23/29.33 & 35.10/40.28
& 4.85/4.17 & 13.77/15.20 & 20.58/22.84 \\
Galileo\cite{tseng2025galileolearningglobal} & 9.78/11.83 & 15.63/17.91 & 27.92/35.28
& 9.95/9.03 & 14.24/15.89 & 22.80/26.16 \\
Skysense\cite{Guo2024} & 13.42/14.47 & 28.54/32.54 & 32.87/36.41
& 8.93/9.38 & 15.29/16.12 & 26.82/31.08 \\
\midrule
UNet Baseline\cite{ronnebergerUNetConvolutionalNetworks2015} & 13.52/16.44 & 25.75/30.08 & 30.16/37.56
& 10.64/11.40 & 14.95/15.10 & 27.00/30.57 \\
ViT Baseline\cite{dosovitskiy2020image} & 12.58/15.72 & 29.43/34.07 & 37.37/42.57
& 10.88/10.03 & 15.26/16.75 & 30.72/31.77 \\
\midrule
Presto\cite{tseng_lightweight_2024} & - & - & - & 14.25 & \underline{27.93} & \underline{34.04} \\
AlphaEarth\cite{SatelliteEmbeddingV1} & \underline{27.12} & \textbf{46.42} & \textbf{51.08} & \underline{14.64} & 21.75 & 25.70 \\
\textbf{TESSERA (Ours)} & \textbf{27.54} &
\underline{46.04} & \underline{50.68} & \textbf{28.20} & \textbf{48.29}
& \textbf{53.12} \\
\bottomrule
\end{tabular}
}
\vspace{6pt}
\hrule
\vspace{6pt}
\textbf{(c) Regression --- Biomassters / Borneo CHM (RMSE
$\downarrow$)}\\[2pt]
\resizebox{\linewidth}{!}{
\begin{tabular}{l|ccc|ccc}
\toprule
\multirow{2}{*}{\textbf{Model}} &
\multicolumn{3}{c|}{\textbf{Biomassters}} &
\multicolumn{3}{c}{\textbf{Borneo CHM}} \\
& 1\% & 30\% & All & 1\% & 30\% &
All \\
\midrule
CROMA\cite{fuller2023croma} & 46.58/44.28 & 39.24/36.80 & 37.14/35.07
& 44.52/41.89 & 31.25/29.67 & 20.56/19.48 \\
DOFA\cite{xiong2024neuralplasticityinspiredmultimodalfoundation} & 51.86/47.84 & 45.72/43.59 & 43.05/41.10
& 43.96/40.68 & 28.97/27.93 & 17.63/17.51 \\
Prithvi\cite{schmude2024prithviwxcfoundationmodel} & 43.89/39.37 & 41.49/38.68 & 41.12/38.88
& 43.50/44.27 & 29.42/27.80 & 19.71/19.41 \\
RemoteCLIP\cite{remoteclip} & 61.70/61.01 & 52.13/51.32 & 49.21/47.92
& 43.24/43.10 & 32.78/30.78 & 28.30/27.59 \\
SatlasNet\cite{Bastani2023} & 49.51/44.64 & 44.24/43.23 & 41.93/41.36
& 45.85/43.01 & 34.01/32.26 & 22.03/21.02 \\
Scale-MAE\cite{Reed2023} & 56.41/51.51 & 51.02/49.34 & 47.05/45.17
& 47.51/46.72 & 32.00/31.11 & 21.12/20.41 \\
SpectralGPT\cite{SpectralGPT} & 64.81/62.28 & 38.54/36.12 & 36.87/36.12
& 43.91/43.65 & 26.03/25.08 & 16.53/15.89 \\
Galileo\cite{tseng2025galileolearningglobal} & 52.72/49.15 & 39.55/36.82 & 38.52/36.92
& 44.82/42.78 & 32.18/31.92 & 20.48/20.27 \\
Skysense\cite{Guo2024} & 47.52/42.28 & 37.51/37.10 & 32.52/30.78
& 42.74/40.96 & 27.42/26.65 & 16.56/\underline{15.58} \\
\midrule
UNet Baseline\cite{ronnebergerUNetConvolutionalNetworks2015} & 44.62/41.70 & 37.92/36.16 & 36.41/35.17
& 41.81/38.61 & 28.01/26.02 & 17.79/16.87 \\
ViT Baseline\cite{dosovitskiy2020image} & 48.24/49.42 & 38.56/38.27 & 37.98/37.52
& 42.02/39.95 & 27.87/26.61 & 16.48/15.77 \\
\midrule
Presto\cite{tseng_lightweight_2024} & - & - & - & 36.43 & 20.95 & 17.88 \\
AlphaEarth\cite{SatelliteEmbeddingV1} & \underline{35.45} & \underline{32.33} & \underline{29.59} & \underline{36.03} & \underline{20.62} &
16.11 \\
\textbf{TESSERA (Ours)} & \textbf{32.50} &
\textbf{29.15} & \textbf{27.43} & \textbf{34.82} & \textbf{15.18}
& \textbf{12.21} \\
\bottomrule
\end{tabular}
}
\end{minipage}
\vspace{-3mm}
\end{table*}

\subsection{Experimental Setup}

\paragraph{Datasets.}
We evaluate \textbf{TESSERA} on six representative benchmarks that span classification, segmentation, and regression tasks. All datasets use multi-temporal Sentinel-1/2 imagery:

\begin{itemize}
    \item \textbf{TreeSatAI-TS}: Tree species classification in Germany.
    \item \textbf{PASTIS-R}: Parcel segmentation in France.
    \item \textbf{Austrian Crop (Classification)}: Crop type classification in Austria~\cite{noauthor_invekos_nodate}.
    \item \textbf{Biomassters}: Above-Ground Biomass (AGB) regression in Finland~\cite{Nascetti}.
    \item \textbf{Borneo Canopy Height}: Canopy height regression in Malaysia~\cite{coomesAirborneLiDARRGB2022}.
    \item \textbf{Austrian Crop (Segmentation)}: Crop semantic segmentation in Austria~\cite{noauthor_invekos_nodate}.
\end{itemize}

To ensure comprehensive evaluation, we select two datasets per task whenever possible: one large-scale (e.g., continental or national coverage) and one fine-grained regional dataset. This design validates both geographic transferability and fine-detail modeling capability. 
High-resolution multi-temporal S1+S2 datasets with dense labels are scarce; hence, we additionally contribute two new benchmarks: \textbf{Austrian Crop} (parcel-level crop mapping) and \textbf{Borneo Canopy Height} (LiDAR-calibrated forest structure in Southeast Asia).

\vspace{-3mm}
\paragraph{Baselines.}
We compare against a comprehensive suite of representative remote-sensing foundation
models (RSFMs): 
CROMA~\cite{fuller2023croma},
DOFA~\cite{xiong2024neuralplasticityinspiredmultimodalfoundation},
Prithvi~\cite{schmude2024prithviwxcfoundationmodel},
RemoteCLIP~\cite{remoteclip},
SatlasNet~\cite{Bastani2023}, 
Scale-MAE~\cite{Reed2023},
SpectralGPT~\cite{SpectralGPT}, 
Galileo~\cite{tseng2025galileolearningglobal},
Skysense~\cite{Guo2024},
Presto~\cite{tseng_lightweight_2024}, and AlphaEarth
~\cite{SatelliteEmbeddingV1}.
We also include standard 
UNet~\cite{ronnebergerUNetConvolutionalNetworks2015} and 
ViT~\cite{dosovitskiy2020image} baselines for reference.

\begin{table}[t]
  \centering
  \caption{\textbf{Model specification.} Multi-temporal and multi-modal (multiple satellite sources) flags, venue, and parameter size. \emph{Params (M)} indicates encoder frozen / fully fine-tuned.}
  \label{tab:model_spec}
  \vspace{-2mm}
  \scriptsize
  \setlength{\tabcolsep}{3pt}
  \renewcommand{\arraystretch}{0.8}
  \begin{tabular}{l >{\centering\arraybackslash}p{12mm} >{\centering\arraybackslash}p{12mm} l c}
    \toprule
    \textbf{Model} & \textbf{Multi-temporal} & \textbf{Multi-modal} & \textbf{Publication} & \textbf{Params (M)} \\
    \midrule
    CROMA       & $\times$ & \checkmark & NeurIPS'23 & 49.1 / 704.9 \\
    DOFA        & $\times$ & \checkmark & arXiv'24   & 40.6 / 151.7 \\
    Prithvi     & \checkmark & \checkmark & arXiv'24 & 40.6 / 126.8 \\
    RemoteCLIP  & $\times$ & \checkmark & TGRS'24   & 40.6 / 128.1 \\
    SatlasNet   & \checkmark & \checkmark & ICCV'23  & 33.3 / 121.2 \\
    Scale-MAE   & $\times$ & $\times$    & ICCV'23  & 49.1 / 352.2 \\
    SpectralGPT & \checkmark & $\times$   & TPAMI'24 & 183.5 / 268.9 \\
    Galileo     & \checkmark & \checkmark & ICML'25  & 39.4 / 125.9 \\
    Skysense    & \checkmark & \checkmark & CVPR'24  & 36.5 / 341.5 \\
    \midrule
    UNet        & $\times$ & $\times$    & MICCAI'15 & 14.8 / 14.8 \\
    ViT-B/16    & $\times$ & $\times$    & ICLR'21   & 125.1 / 125.1 \\
    \midrule
    Presto      & \checkmark & \checkmark & arXiv'23 & 30.2 / 30.2 \\
    AlphaEarth  & \checkmark & \checkmark & arXiv'25 & 30.1 / 30.1 \\
    \textbf{TESSERA (Ours)} & \checkmark & \checkmark & -- & 30.2 / 30.2 \\
    \bottomrule
  \end{tabular}
  \vspace{-6.5mm}
\end{table}

\vspace{-3mm}
\paragraph{Implementation.}
For embedding-based models (TESSERA, Presto, AlphaEarth), we train a 2-layer MLP for pixel-wise classification, and a lightweight UNet for patch-based classification, segmentation, and regression. 
Patch-based models are fine-tuned using UPerNet heads. 
All models are evaluated at 1\%, 30\% and 100\% label ratios to study label efficiency. 
Additional implementation details are in the supplementary material.

\subsection{Classification Results}

\paragraph{Patch-based classification: TreeSatAI-TS.}
\Cref{tab:unified_result_table} (a1) reports results on
TreeSatAI-TS, a national-scale forest species benchmark. 
\textbf{TESSERA} achieves a state-of-the-art F1-score of \textbf{77.96} with full supervision, significantly outperforming all other models, including the next-best competitor, AlphaEarth (76.90)\footnote{The TESSERA encoder (45.7M parameters) used during these tasks is also significantly smaller than AlphaEarth's (480M parameters).} and the strongest fine-tuned baseline, the UNet Baseline (73.30).
This highlights how our pixel-level temporal modeling effectively
captures phenological variations, which improve class separability \cite{immitzer2019optimal}.
Notably, TESSERA's superiority is even more pronounced in the few-shot setting: at 1\% labels, it achieves an F1 score of \textbf{60.58} ,
surpassing the second-best model, AlphaEarth (52.79), by nearly 8 points.

\begin{figure}[t]
    \centering
    \includegraphics[width=1\linewidth]{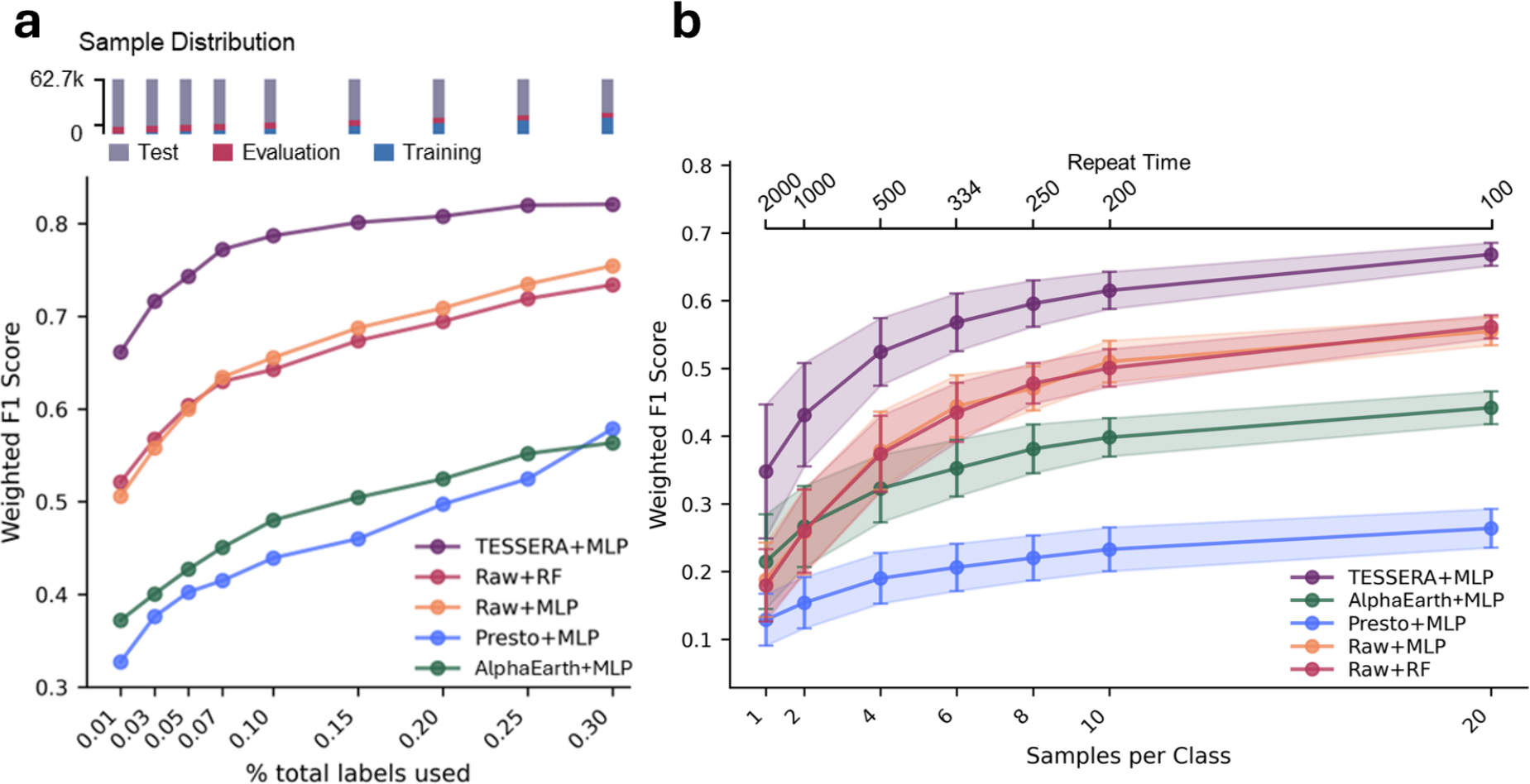}
    \caption{\textbf{Label-efficient crop classification on Austrian Crop.}
    \textbf{(a)} Weighted F1 vs. training label ratio using a small head on frozen embeddings.
    \textbf{(b)} Few-shot performance. TESSERA attains strong accuracy with very few labels; error bars denote variation over runs.}
    \label{fig:austrian_crop}
    \vspace{-5mm}
\end{figure}

\vspace{-3mm}
\paragraph{Pixel-wise crop classification: Austrian Crop.}
On the \textbf{Austrian Crop} dataset
(\Cref{tab:unified_result_table} (a2)), 
TESSERA demonstrates strong performance. 
As visualized in~\Cref{fig:austrian_crop} (a),
it consistently surpasses Presto and AlphaEarth in all standard label ratios (1\%-30\%).
The performance gap is most pronounced in the low-data regime: at just 1\%
labeled data, TESSERA achieves a \textbf{66.15} F1 score.
This outperforms AlphaEarth (37.22) by more than 28.9
points and Presto (32.74) by more than 33.4 points.
We further investigate extreme label scarcity in~\Cref{fig:austrian_crop} (b), evaluating performance with only 1 to 20 labeled samples per class. 
In this challenging few-shot scenario, TESSERA's performance
remains robust and scales gracefully (e.g., $\sim$0.5 F1 with
just 4 samples/class), while the F1 scores of other models are below 0.4.

\vspace{-2mm}
\subsection{Segmentation Results}

\paragraph{Field parcel segmentation: PASTIS-R.}
On the large-scale PASTIS-R benchmark
(\Cref{tab:unified_result_table} (b)), 
TESSERA achieves competitive performance. With full labels,
it obtains \textbf{50.68} mIoU, ranking second
and just behind AlphaEarth (51.08). 
However, in the low-label regime (1\% labels), TESSERA (\textbf{27.54}) surpasses AlphaEarth (27.12) and significantly outperforms all fine-tuned baselines, such as the UNet Baseline (16.44) and Skysense (14.47).
Unlike some RSFMs that require compute-heavy decoders ($>$100M parameters), our compact 30M-parameter UNet suffices. This demonstrates that spatial context can be learned efficiently using TESSERA's pre-trained temporal embeddings.

\vspace{-3mm}
\paragraph{Regional crop segmentation: Austrian Crop.}
For the fine-grained Austrian Crop segmentation task, TESSERA achieves \textbf{53.12} mIoU, establishing a new SOTA and significantly improving over both AlphaEarth (25.70) by +27.4 points and Presto (34.04) by +19.0 points. 
This performance gain is consistent across all label regimes, especially at 1\% labels (TESSERA: \textbf{28.20} vs AlphaEarth: 14.64), confirming the effective transferability of temporally coherent embeddings. 
TESSERA yields sharper parcel edges and reduces confusion
among visually similar crops (see supplementary materials), indicating its ability to maintain semantic consistency over time.

\subsection{Regression Results}

\begin{figure}[t]
  \centering
  \includegraphics[width=\linewidth]{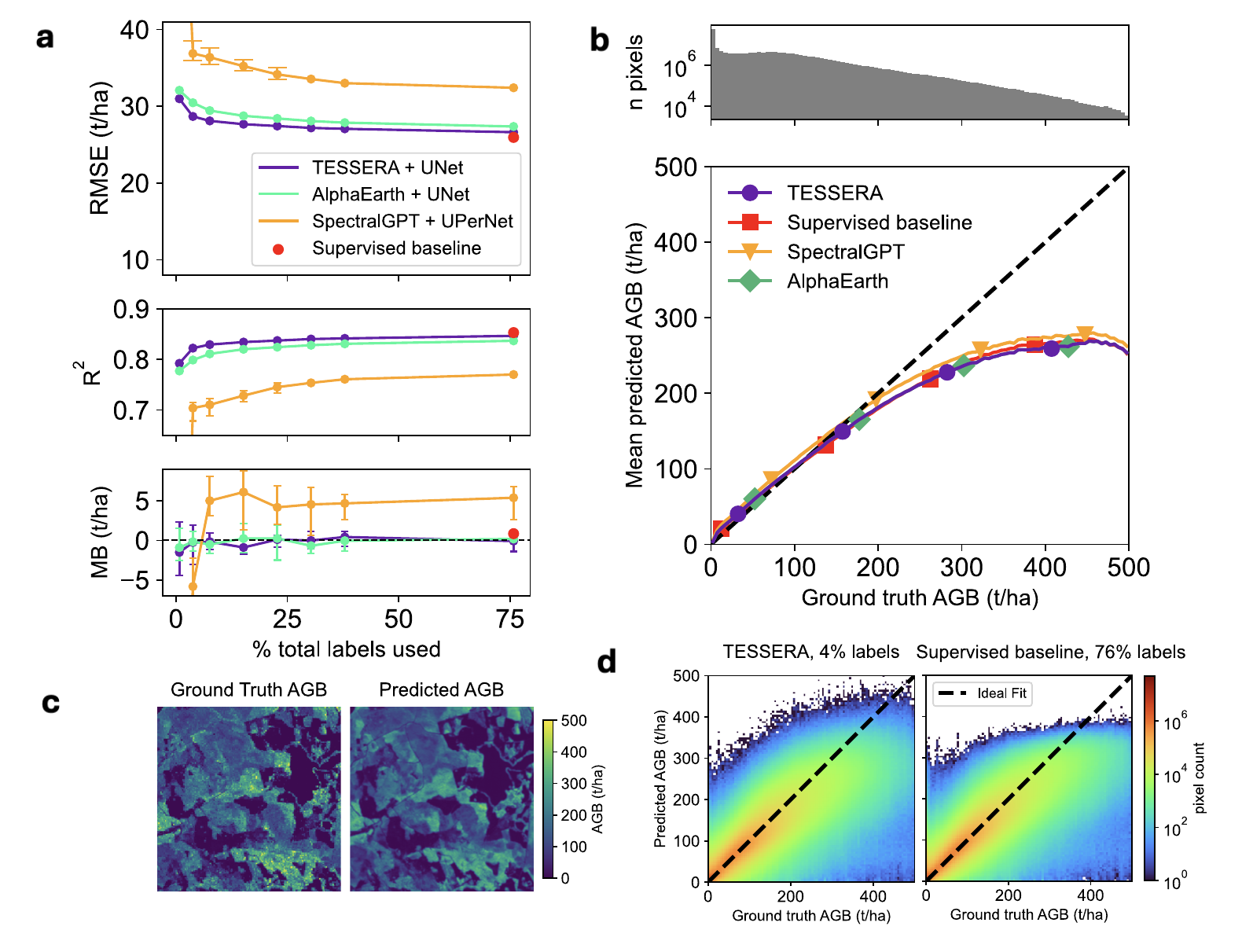}
  \caption{\textbf{Biomassters AGB regression.}
  \textbf{(a)} RMSE/$R^2$/MB (mean bias) vs. label fraction.
  \textbf{(b)} Predicted AGB vs.~Ground truth AGB.
  \textbf{(c)} Spatial AGB maps from TESSERA trained with 4\% labels.
  \textbf{(d)} Predicted vs.\ ground-truth scatter.
  TESSERA closely tracks the task-specific winner with significantly fewer labels.}
  \label{fig:agb}
\end{figure}

\paragraph{Above-ground biomass estimation: Biomassters.}
As shown in~\Cref{tab:unified_result_table} (c),
TESSERA achieves an RMSE of \textbf{27.43} t/ha with full labels -- outperforming all other foundation models, including the next-best, AlphaEarth (29.59 t/ha). 
This result surpasses the strongest fine-tuned RSFM baseline,
Skysense (30.78 t/ha).
The results demonstrate that temporal embeddings capture biomass dynamics effectively, even under limited supervision. 
\Cref{fig:agb} provides a detailed comparison for this task after clipping implausibly high biomass values above 500 t/ha to this threshold.
In this analysis, TESSERA (\textbf{26.61} t/ha) nearly matches the best-performing supervised task-specific model (25.90 t/ha), which won the Biomassters competition of more than 1000 submissions.
The figure further illustrates that TESSERA maintains consistent performance in all label ratios, shows minimal bias in all biomass ranges, and generates spatially coherent AGB maps with only 4\% of total labels.

\vspace{-5mm}
\paragraph{Canopy height regression: Borneo Canopy Height.}
In the Borneo CHM dataset, TESSERA again demonstrates
superior performance, achieving the lowest RMSE of \textbf{12.2} m.
This result significantly outperforms all other models,
including AlphaEarth (16.1 m), Presto (17.9 m), and the best-performing
fine-tuned RSFM such as Skysense (15.6 m) and the ViT Baseline (15.8 m).
We also compare with DINOv2-derived and Sentinel-2 composite
height maps, where TESSERA demonstrates the closest agreement with LiDAR ground truth (see supplementary materials).

\section{Discussion \& Ablation Studies}
\label{sec:ablation}

\begin{figure*}[t]
    \centering
    \includegraphics[width=\linewidth]{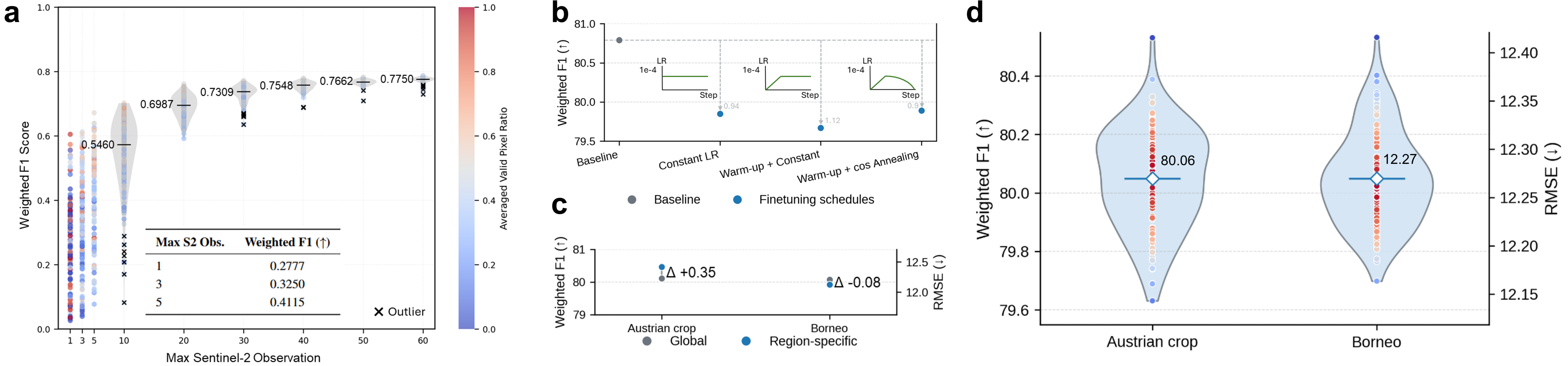}
    \caption{\textbf{Robustness and sensitivity analyses.}
    \textbf{(a)} \textit{Cloud robustness:} performance remains stable until Sentinel-2 valid observations per year drop below $\sim$20, after which accuracy degrades markedly.
    \textbf{(b)} \textit{Finetuning impact:} evaluating the effect of three different learning rate strategies during finetuning.
    \textbf{(c)} \textit{Embedding specificity:} comparing the performance of a global embedding versus an embedding retrained on region-specific data.
    \textbf{(d)} \textit{Temporal sampling uncertainty:} selecting $L{=}40$ timesteps within a year produces consistently similar results across random choices.}
    \label{fig:compostite_graph_discussion}
    \vspace{-5mm}
\end{figure*}

\begin{figure}[t]
    \centering
    \includegraphics[width=\linewidth]{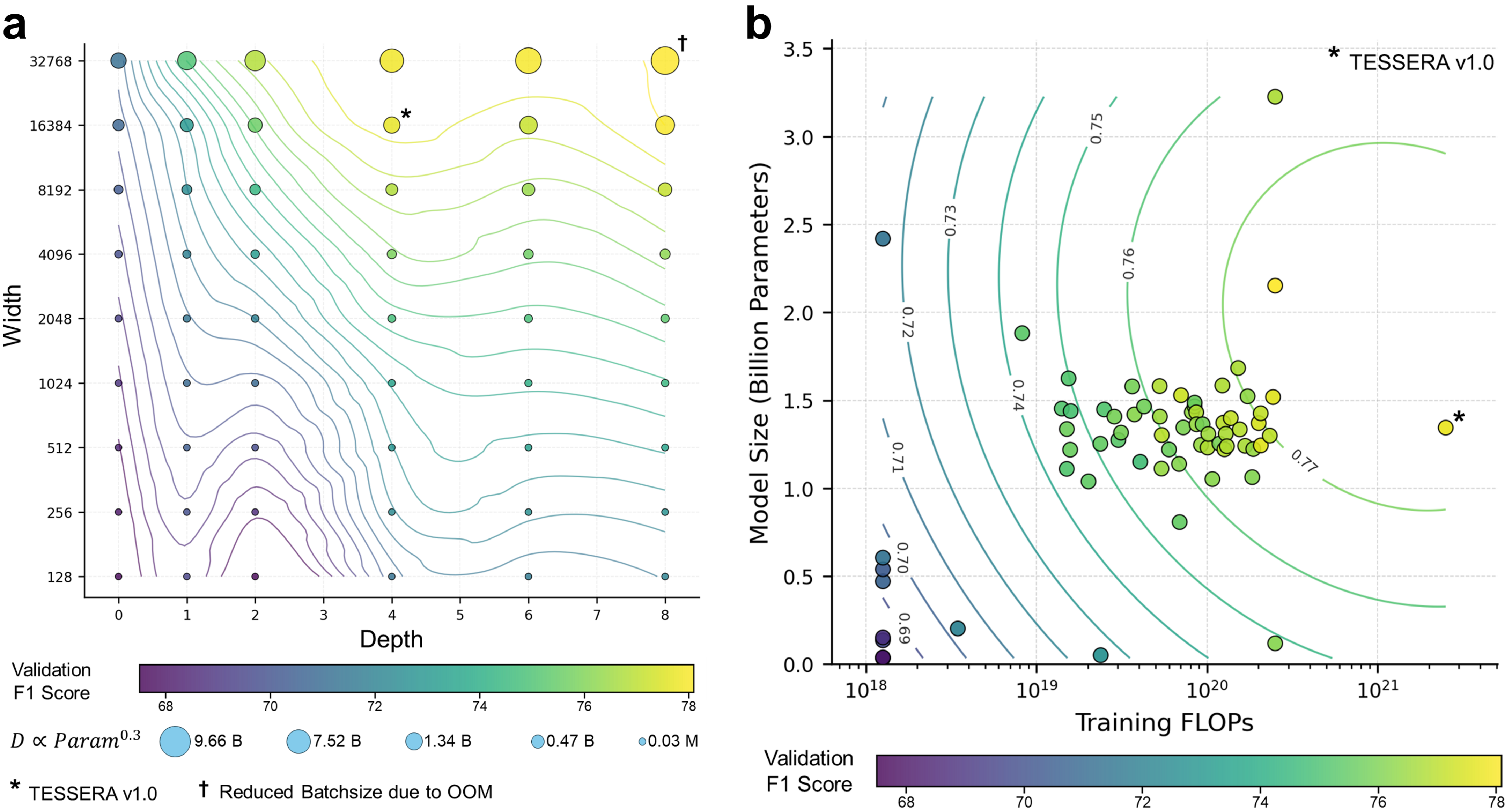}
    \caption{\textbf{Scaling behavior.}
    \textbf{(a)} Projector depth/width sweep: validation F1 improves with capacity; † marks the max-F1 setup under memory constraints; * denotes the chosen \textbf{TESSERA v1.0}.
    \textbf{(b)} Validation F1 scales predictably with training FLOPs and parameters, with \textbf{v1.0} near the compute-accuracy frontier.}
    \label{fig:scaling_laws}
    \vspace{-3mm}
\end{figure}

We validate our design choices on the Austrian Crop classification task. \Cref{tab:ablation} summarizes the results; in the following we organize the findings with brief Q\&A headings and defer all methodological details to the supplementary material.

\begin{table}[t]
\centering
\caption{\textbf{Ablation on the Austrian Crop dataset.} We
report \textbf{Validation F1} and \textbf{RankMe} (higher $\uparrow$ is
better). Parentheses show change vs.~the Baseline.}
\label{tab:ablation}
\vspace{-2mm}
\setlength{\tabcolsep}{6pt}
\renewcommand{\arraystretch}{0.8}
\resizebox{\columnwidth}{!}{%
\begin{tabular}{lcc}
\toprule
\textbf{Configuration} & \textbf{Val. F1 ($\uparrow$)}
& \textbf{RankMe ($\uparrow$)} \\
\midrule
\textbf{Baseline (Full Model)} & \textbf{77.3} &
\textbf{0.963} \\
\midrule
w/o Global Data Shuffling  & 68.1 (\textcolor{red}{-9.2}) & 0.847 (\textcolor{red}{-0.116}) \\
w/o Mixup Regularization   & 66.2 (\textcolor{red}{-11.1}) & 0.857
(\textcolor{red}{-0.106}) \\
w/o Sentinel-1 Data        & 74.2 (\textcolor{red}{-3.1}) & 0.931 (\textcolor{red}{-0.032}) \\
w/o Shuffling \& Mixup     & 62.6 (\textcolor{red}{-14.7}) &
0.867 (\textcolor{red}{-0.096}) \\
w/o Embedding Quantization & 77.9 (\textcolor{green}{+0.6}) & 0.972
(\textcolor{green}{+0.009}) \\
\midrule
w/o Pretraining & 43.8(\textcolor{red}{-33.5}) &  - \\
\bottomrule
\end{tabular}%
}
\vspace{-2mm}
\end{table}

\vspace{-3mm}
\paragraph{Do shuffling and mixup help?}
Yes. Removing global shuffling or mixup drops validation F1 by $-9.2$/$-11.1$ and RankMe score by $-0.116$/$-0.106$. Removing both leads to the largest fall ($-14.7$ F1; see~\Cref{tab:ablation}). The two strategies are complementary: shuffling smoothes the loss curve, mitigates spatial autocorrelation while mixup helps to prevent feature overfitting~\cite{Bandara2023MixCoAM}. 

\vspace{-3mm}
\paragraph{How robust are the embeddings to cloudiness?}
We find that when the annual number of non-cloudy \emph{valid} Sentinel-2 observations drops $\leq\!10$, Macro-F1 performance drops sharply (see~\Cref{fig:compostite_graph_discussion}). At a global scale, we find that embeddings are quite robust to cloudiness since there are more than 70 observations per year for each orbit.

\vspace{-3mm}
\paragraph{Is fine-tuning weights necessary?}
Self-supervised pretraining of the encoder yields large performance gains compared to randomly initialized weights, a +33.5 F1 improvement in the Austrian Crop classification task. Pretraining, somewhat surprisingly, makes subsequent finetuning unnecessary: \Cref{fig:compostite_graph_discussion} (b) shows that finetuning the encoder provides no benefit. This suggests that the embeddings encapsulate a highly generalized representation.

\vspace{-3mm}
\paragraph{Are regional embeddings necessary?} We find that retraining with region-specific data offers negligible gains (\Cref{fig:compostite_graph_discussion} c). A single global model is sufficient for high performance across diverse geographies, reinforcing the \emph{Embeddings-as-Data} paradigm and making costly regional retraining unnecessary.

\vspace{-3mm}
\paragraph{Can a spatial head recover spatial context?}
Yes. Although pretraining is pixel-wise (with no explicit spatial structure), adding a spatial head (e.g., a UNet) consistently improves segmentation and regression tasks, with larger gains at higher label budgets. This suggests that spatial cues can be injected \emph{downstream} efficiently without complicating pretraining or creating purely pixel-based embeddings.

\vspace{-3mm}
\paragraph{How sensitive are embeddings to random temporal sampling?}
Not much. With $L{=}40$ temporal samples, repeated random draws produce tightly clustered results (low variance), consistent with our time-sampling invariance objective (shown in \Cref{fig:compostite_graph_discussion} c). We also compare $L\!\in\!\{20,40,96,365\}$ and find that $L{=}40$ is the accuracy–efficiency sweet spot (see supplementary material). 

\vspace{-3mm}
\paragraph{Scaling laws.}
\Cref{fig:scaling_laws} shows that the performance scales predictably with the depth/width of the projector, FLOPs, and the parameters; TESSERA v1.0 (star) lies near the optimal compute–accuracy frontier, while the dagger~(†) marks the maximum-F1 configuration constrained by memory/batch size.

\vspace{-3mm}

\paragraph{Flexibility of the inference time window.}
In the default setting, TESSERA is trained using a one-year temporal window.
However, this one-year window is not a strict requirement.
We find that applying the model to shorter seasonal windows during inference (e.g., only from spring to autumn) still yields effective embeddings
(see supplementary materials for seasonal window analyses on the Austrian crop dataset).

\vspace{-3mm}
\paragraph{Does \texttt{int8} quantization significantly degrade performance?}
No: there is only a small drop in performance. Removing quantization slightly \emph{improves} metrics (e.g., $+0.6$ F1 and $+0.009$ RankMe in~\Cref{tab:ablation}), indicating that \texttt{int8} introduces minor information loss. However, using \texttt{int8} compresses the embeddings to $\sim\!25\%$ of the \texttt{fp32} storage footprint (and reduces I/O), which is attractive for global-scale products and for use on an edge device.

\section{Conclusion}
\label{sec:conclusion}

We introduced TESSERA, a pixel-wise foundation model for Earth Observation that learns robust representations from sparse, multi-modal time series via temporal sampling invariance. Extensive experiments show that TESSERA establishes a new state-of-the-art across diverse tasks with unparalleled label efficiency. We pioneer the ``Embeddings-as-Data'' paradigm, releasing analysis-ready global embeddings via the \texttt{GeoTessera} library to lower barriers for geospatial analysis and empower future research.

\section{Acknowledgments}
\label{sec:ack}

We thank UKRI (MR/Z505456/1), DAWN, AMD, Vultr, the STFC Durham DiRAC HPC Facility (ST/P002293/1, ST/R002371/1, ST/S002502/1, ST/R000832/1) and the Microsoft AI For Good Lab for resources. Donations from Tarides, Jane Street, Google, Dr. Robert Sansom and John Bernstein also financially supported this research.
{
    \small
    \bibliographystyle{ieeenat_fullname}
    \bibliography{main}
}

\clearpage
\appendix

\section{Supplementary Material}
\label{sec:suppl}

In this supplementary material, we provide additional details on downstream tasks, model architecture, and training procedures that were omitted from the main paper due to space constraints.

\subsection{Taxonomy of Remote Sensing Foundation Models}
\begin{figure*}[!h]
    \centering
   \includegraphics[width=1\linewidth]{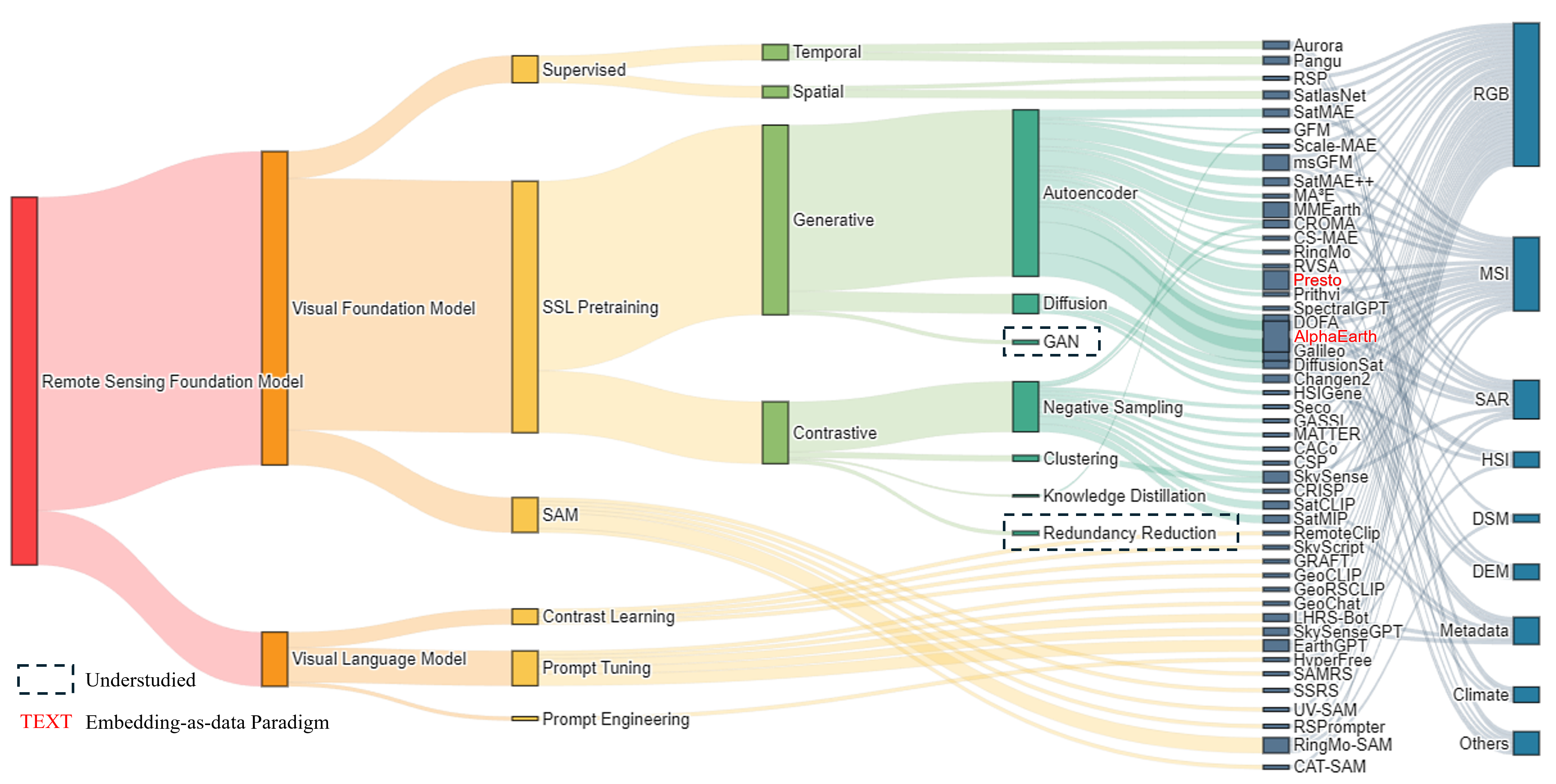}
    \caption{Taxonomy of RSFMs. All RSFMs are categorized into two primary classes: visual foundation models and visual language models, which are further subdivided according to different methodologies. The final column of the Sankey diagram illustrates the data modalities used by each foundation model. Dashed rectangles indicate areas with minimal or virtually no existing research. The checkerboard pattern denotes that, in addition to providing model parameters, these foundation models also offer globally generated, analysis-ready representation products derived from the model itself.}
    \label{fig:taxonomy}
\end{figure*}

The current landscape of RSFMs, as illustrated in ~\Cref{fig:taxonomy}, is vibrant and rapidly expanding, primarily along two avenues: Visual Foundation Models (VFMs) and Visual Language Models (VLMs). While these models have pushed the boundaries of what is possible in terms of performance, they have largely overlooked the critical dimension of usability, creating a significant bottleneck for the broader scientific community.

VFMs are engineered to interpret a diverse array of visual remote sensing data. Early and influential models like SatMAE~\cite{Cong2022} and its successors have successfully adapted the masked autoencoder (MAE)~\cite{he2022masked} paradigm for temporal MSI and RGB imagery, demonstrating strong capabilities in learning spatial patterns. For instance, Scale-MAE~\cite{Reed2023} introduced a scale-aware pre-training strategy to handle multi-scale geospatial data, while SatMAE++~\cite{Noman2024} explored multi-scale reconstructions to better capture scale variability in RS images. To address the need for multi-modal fusion, models like MMEarth~\cite{Nedungadi2024} and the recently proposed Galileo~\cite{tseng2025galileolearningglobal}
have emerged, integrating an extensive range of inputs including MSI, SAR, and DEM through sophisticated fusion frameworks. Concurrently, models focused on temporal analysis have also been developed. Presto~\cite{tseng_lightweight_2024}, for instance, is a pixel-level Transformer pre-trained on a global time-series dataset, showing excellent performance in tasks requiring temporal understanding. 

However, a unifying weakness across almost all these VFMs is their profound usability challenge. They are released as massive models, placing the onus of data acquisition, complex preprocessing, and computationally intensive fine-tuning squarely on the end-user. This approach requires users to possess significant GPU resources and deep technical expertise, creating a high barrier to entry. While Google's AlphaEarth makes a notable exception by providing pre-generated representation maps, it is not fully open-source.

VLMs for remote sensing aim to bridge the gap between visual data and natural language, partially addressing cross-domain integration. Models like SatCLIP ~\cite{klemmer2025satclip}, RemoteClip ~\cite{remoteclip}, and GeoRSCLIP~\cite{zhang2024rs5m} have adapted contrastive learning (e.g., CLIP~\cite{radford2021learning}) to align satellite images with textual descriptions, enabling powerful zero-shot classification and text-based image retrieval. More advanced models like EarthGPT~\cite{zhang2024earthgpt} and SkySenseGPT~\cite{luo2024skysensegpt} build upon frameworks like LLaVA~\cite{liu2023visual} to facilitate conversational interaction and multi-sensor (e.g., SAR, infrared) image comprehension. While these models open up exciting new avenues for human-computer interaction with geospatial data, they inherit the same fundamental usability limitations as VFMs. They are typically released as large models requiring expert handling and significant computational power, and none provide large-scale, pre-generated products that would lower the barrier to entry for non-specialist users.

In summary, the field has made tremendous strides in improving the \textit{performance} of RSFMs. Yet, this progress has been largely confined to a model-centric paradigm, inadvertently creating a high barrier to entry that contradicts the goal of democratizing science. The critical gap, therefore, is not a lack of powerful models, but a lack of a user-centric framework that delivers their power in an accessible and ready-to-use format.

\subsection{Model Details}
\paragraph{Dual-Encoder Architecture}
Given the distinct nature of Sentinel-1 SAR and Sentinel-2 MSI data, TESSERA employs two separate, parallel transformer-based encoder branches.
\begin{itemize}
    \item \textbf{Sentinel-2 MSI Encoder}: This branch processes a time series of 10 spectral bands (excluding the 60~m bands used for the detection of water vapour and clouds) from Sentinel-2. We used blue (B2), green (B3), red (B4), red edges 1--3 (B5, B6, B7), near-infrared (B8, B8A), and shortwave infrared (B11, B12).
    \item \textbf{Sentinel-1 SAR Encoder}: This branch processes a time series of 2 polarizations from Sentinel-1 (VV and VH).
\end{itemize}
Each encoder begins by linearly embedding the input features (spectral bands or polarizations) for each time step. To preserve sequence order and incorporate temporal context, learnable positional encodings based on the Day-of-Year (DOY) of each observation are added to these embeddings. The core of each encoder consists of a stack of 4 standard Transformer blocks~\cite{Vaswani2017AttentionIA}, featuring multi-head self-attention and feed-forward layers to learn temporal patterns within the data streams.

To derive a single vector summarizing the entire time series for each modality, an attention-pooling layer weighs the importance of different time steps before aggregation. The resulting modality-specific embeddings (one from the S1 encoder, one from the S2 encoder) are then fused using a multi-layer perceptron.

\paragraph{Projector Network}
\label{sec:projector}
The fused embedding from the dual-encoder stage is subsequently fed into a large projector network. This projector is a six-layer MLP. Its architecture comprises an input layer mapping the fused embeddings to 16,384 dimensions, followed by four hidden layers of 16,384 dimensions each, and a final linear output layer. Each of the first five layers is a fully-connected linear layer followed by Batch Normalization and a ReLU non-linear activation function, making the network deep and highly non-linear. This significant expansion in dimensionality is crucial, as suggested by the original Barlow Twins work~\cite{Zbontar2021BarlowTS}, to enable effective redundancy reduction during the self-supervised loss computation. The final output of the projector is then used in the loss calculation. For downstream tasks, we used the 128-dimensional output from the fusion MLP (before the projector) as the final pixel embedding. The TESSERA encoder (up to the fusion MLP) has approximately 46 million parameters, while the projector accounts for the majority of the model's $\sim$1.4 billion parameters.

\begin{table}[!h]
\centering
\caption{\textbf{Architectural parameters of the TESSERA v1.0 model.}}
\label{tab:param_breakdown}

\setlength{\tabcolsep}{4pt}
\renewcommand{\arraystretch}{0.95}

\resizebox{\linewidth}{!}{%
\begin{tabular}{lrr}
\hline
\textbf{Component} & \textbf{Parameters} & \textbf{Percentage of Total} \\
\hline
\textbf{Encoder (Feature Extractor)} & \textbf{45,697,026} & \textbf{3.29\%} \\
\quad Sentinel-1 Encoder & 22,846,465 & 1.64\% \\
\quad Sentinel-2 Encoder & 22,850,561 & 1.64\% \\
\\
\textbf{Fusion MLP} & \textbf{131,200} & \textbf{$<$0.01\%} \\
\\
\textbf{Projector (for Training only)} & \textbf{1,344,700,421} & \textbf{96.71\%} \\
\hline
\textbf{Total Parameters} & \textbf{1,390,528,647} & \textbf{100.00\%} \\
\hline
\end{tabular}
} 
\end{table}

\paragraph{Self-Supervised Training}
\label{app:ssl_training}

The TESSERA model is trained using a modified Barlow Twins objective function~\cite{Zbontar2021BarlowTS}. For this objective, two distorted views, denoted as $Y_A$ and $Y_B$, are generated for each input d-pixel. In TESSERA, these views are created by independently running the temporal sampling and preprocessing pipeline twice for the Sentinel-1 and Sentinel-2 data associated with a given d-pixel. This process involves:
\begin{enumerate}
    \item For each view, independent sampling of a fixed number of valid observation dates from the annual Sentinel-2 time series (10 spectral bands).
    \item For each view, independent sampling of a fixed number of valid observation dates from the annual Sentinel-1 time series (2 polarizations).
\end{enumerate}
These views represent different, valid, but inherently incomplete glimpses of the pixel's true temporal-spectral evolution, akin to observing the same location through intermittent cloud cover or from different satellite passes at different times. The model learns by reconciling these partial views. The inherent differences between the Sentinel-1 SAR and Sentinel-2 MSI modalities further provide diverse perspectives on the same underlying physical processes. Thus, our augmentations are fundamentally about sampling from the available, inherently incomplete information streams, rather than artificially distorting a complete input.

\subsection{Pre-training Details}
\label{app:pre-training_details}
The TESSERA model was pre-trained using approximately 800 million d-pixels drawn from 3,012 globally distributed Military Grid Reference System (MGRS) tiles chosen from the years 2017-2024. d-pixels were generated by spatially downsampling Sentinel-1 and Sentinel-2 data by a factor of 400. This was achieved by systematically taking every 20th pixel along the horizontal and vertical axes. Each d-pixel contained the annual time series for 10 spectral bands of Sentinel-2 data and the annual time series of VV and VH polarizations for Sentinel-1 data.

For each pixel location, after removing invalid observations (e.g., due to cloud cover for Sentinel-2), we performed sparse temporal sampling. This involves randomly selecting a fixed number of 40 valid observation dates from the year's data. The sequence length for the Transformer encoders was fixed at 40 timesteps for both modalities. If there are fewer than 40 cloud-free dates, replacement sampling is performed. This strategy standardizes the input sequence length and serves as a key data augmentation mechanism, building invariance to data gaps and teaching the model that the underlying signal persists regardless of the specific dates observed. Finally, these values were standardized using global statistics to stabilize training. The temporal context of each sampled observation was encoded by transforming its normalized Day-of-Year (DOY) into sine and cosine features, which were then concatenated with the corresponding spectral or backscatter measurements. 

\begin{figure*}[t]
    \centering
    \includegraphics[width=1\linewidth]{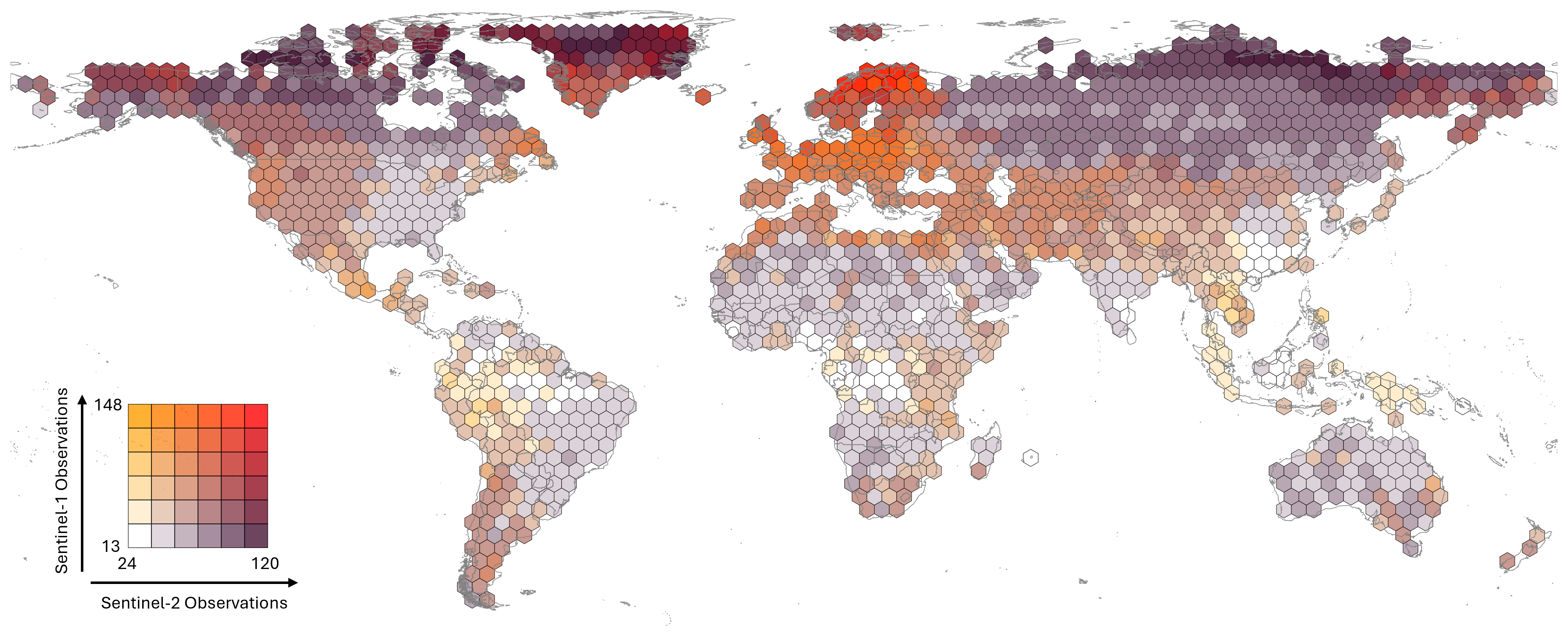}
    \caption{\textbf{TESSERA is trained on globally-distributed training data.} TESSERA was trained on over 3000 MGRS tiles distributed globally from 2017 to 2024. The colour of each hexagon in the map corresponds to the average number of valid observation days for Sentinel-1 (S1) and Sentinel-2 (S2), as defined by the bivariate colour legend. This visualization highlights the density of combined S1 and S2 observations available for training across different regions, ensuring the model learns from a diverse range of geographical and environmental conditions.}
    \label{fig:training_data_distribution}
\end{figure*}

As we were in a compute-limited rather than data-limited pre-training regime, we trained for a single epoch, which corresponded to approximately 6200 GPU hours on 16 AMD MI300X GPUs (192GB memory each). We used PyTorch with Fully Sharded Data Parallel (FSDP) and Automatic Mixed Precision (AMP) enabled.  AdamW optimizer was used with a base learning rate of 0.002 and weight decay of $1 \times 10^{-6}$. The learning rate schedule consisted of a linear warmup during the initial 10\% of steps, followed immediately by a cosine decay for the remainder of training. The global batch size was 32,768. Key training dynamics, including loss curves, learning rate schedule, and evolution of downstream performance during pre-training, are visualized in Supplementary ~\Cref{fig:training_log}.

\begin{figure}[t]
    \centering
    \includegraphics[width=\linewidth]{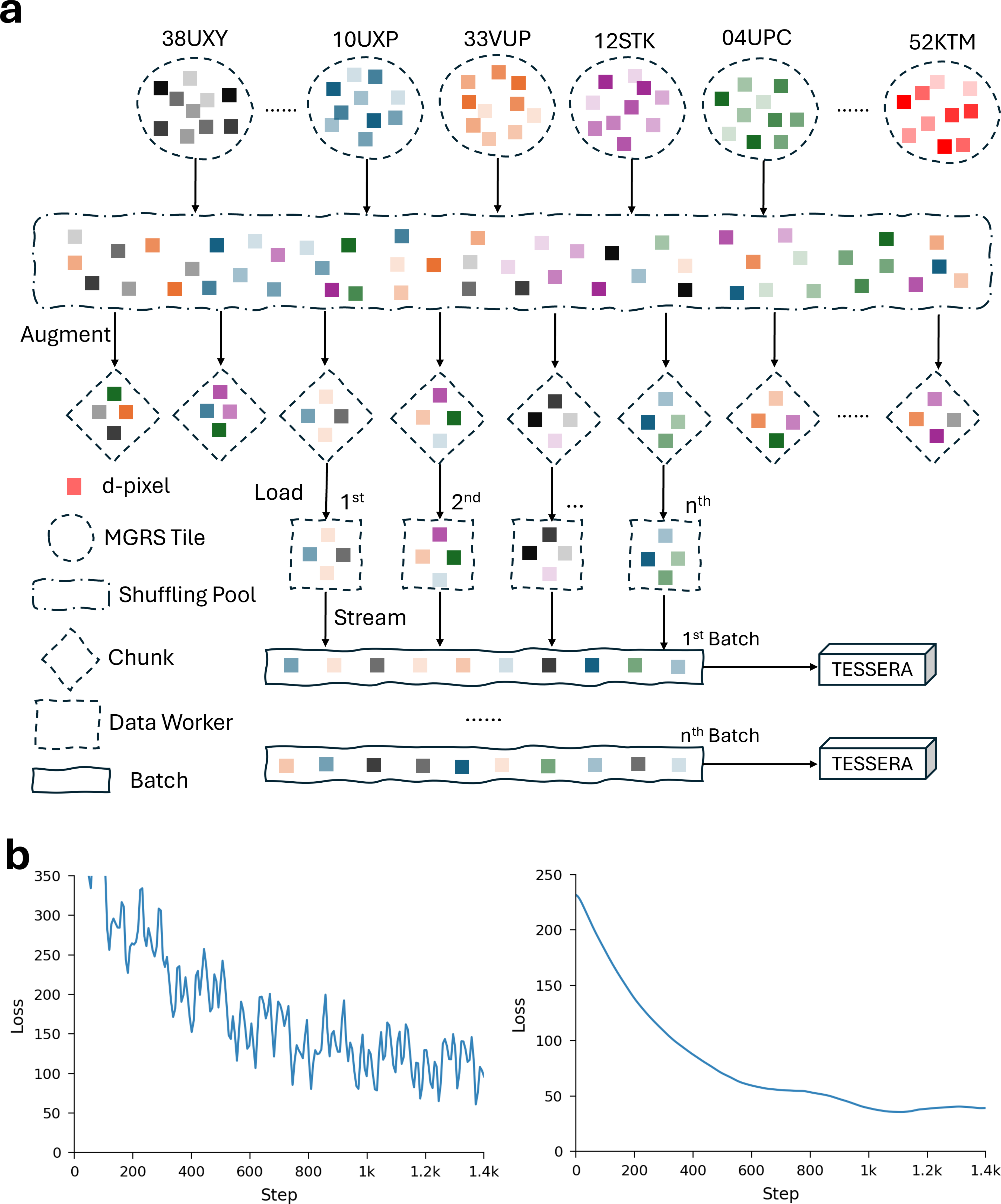}
    \caption{\textbf{Data shuffling greatly improves training stability.} \textbf{a}, Schematic of the data shuffling and loading process. D-pixels (coloured squares) from thousands of MGRS tiles are first aggregated into a global pool. A custom Rust binary performs a global shuffle on this multi-terabyte dataset before applying augmentations. The processed data is then organized into chunks and streamed by data workers to form well-shuffled, globally diverse training batches. \textbf{b}, Comparison of training loss curves. The top plot shows the volatile loss progression typical of a localized shuffling strategy, which is susceptible to geographic bias. The bottom plot shows the significantly smoother and more stable loss curve achieved with our global shuffling pipeline, demonstrating more effective and robust model convergence.}
    \label{fig:shuffle}
\end{figure}

A crucial aspect of our training methodology is a data shuffling strategy, essential for learning globally representative features from a representative sample of a vast and geographically diverse dataset. Given that d-pixels within an individual MGRS tile exhibit high spatial autocorrelation, a naive sequential or locally-shuffled data loading process would expose the model to strong geographic biases in each batch. To overcome this, we developed a custom data processing pipeline to implement a truly global shuffle across all $\approx$800 million training samples, which constituted more than 2~TB of initial d-pixel data. This pipeline is conceptually illustrated in \Cref{fig:shuffle} a.

The impact of this approach on training stability is empirically demonstrated in ~\Cref{fig:shuffle} b. Compared to a conventional localized shuffling strategy, which results in a highly volatile loss curve (top plot), our global shuffling strategy yields a markedly smoother and more stable convergence (bottom plot). This enhanced stability is fundamental for robust convergence and for preventing the model from overfitting to regional characteristics.

\begin{figure*}[t]
    \centering
    \includegraphics[width=\linewidth]{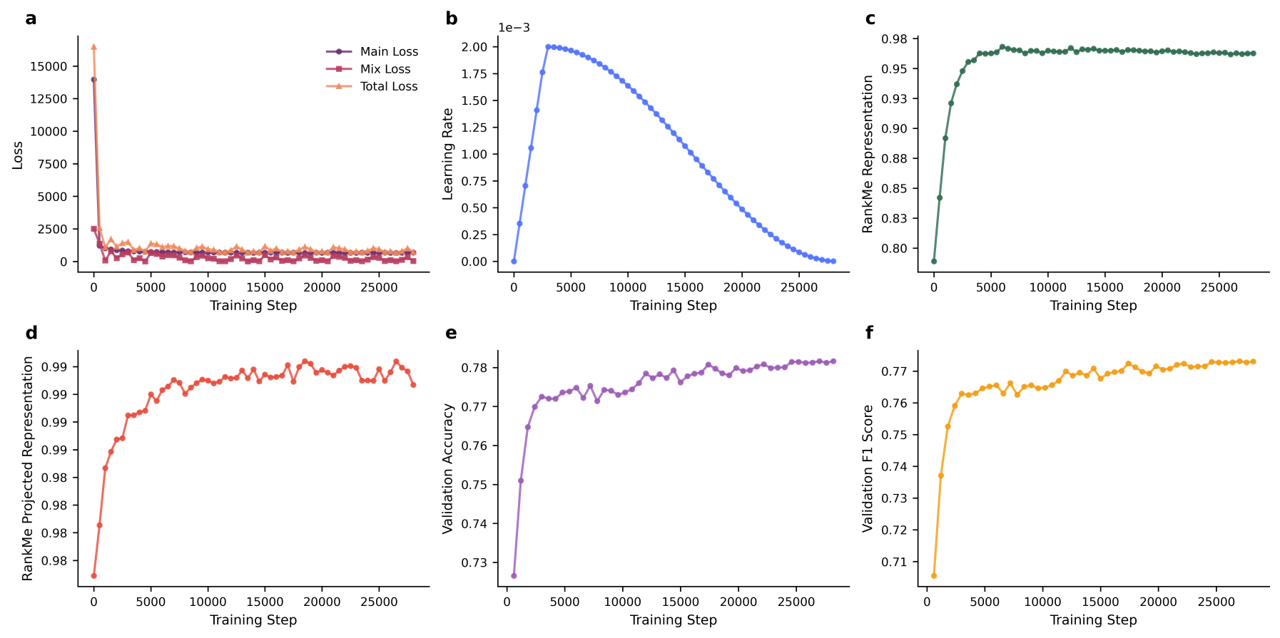}
    \caption{\textbf{Training dynamics curves show that TESSERA rapidly reaches its asymptotes.} 
    \textbf{a}, Training loss curves, showing the main Barlow Twins loss, the mixup regularization loss, and the total loss. 
    \textbf{b}, The learning rate schedule, consisting of a linear warmup phase followed by a cosine decay. 
    \textbf{c}, RankMe score calculated on the high-dimensional (16,384-D) output of the projector, which quickly saturates, indicating the model is effectively utilizing the large feature space for redundancy reduction. 
    \textbf{d}, RankMe score of the final 128-D fused representation, showing that the embeddings used for downstream tasks become progressively richer and less redundant. 
    \textbf{e}, \textbf{f}, Downstream performance on the Austrian crop classification validation set, tracked by accuracy and F1 score respectively during pre-training, showing a steady improvement that correlates with the training progress.}
    \label{fig:training_log}
\end{figure*}

Operationally, the process begins with the aggregation of d-pixels from all MGRS tiles into a single, comprehensive pool. A global shuffling operation is performed on this pool, a critical step to break the spatial contiguity of data from individual tiles and ensure each training batch contains a diverse mix of geographical and environmental contexts. Following this global shuffle, the data augmentation required by the Barlow Twins framework is applied.

To manage the significant I/O demands of shuffling and augmenting such a large volume of data, we developed a custom high-performance pipeline that handles the reading of raw d-pixel data, executes the global shuffle, and prepares the data for augmentation. The resulting pairs of augmented d-pixels are then serialized into a compact, pickle-like file format. These files are organized into manageable chunks and loaded by PyTorch \textit{DataLoader} workers, which stream the data and assemble the final training batches. This end-to-end pipeline ensures that each batch presented to the model is a well-shuffled, globally diverse representation of Earth's surface characteristics, which is fundamental for training a robust pixel-wise foundation model like TESSERA.

\subsection{Global Embedding Map Generation}
\label{app:inference_maps}
A primary output of the TESSERA project is the generation of annual global embedding maps with 10~meter resolution for the years 2017-2024.
To generate these maps, the pre-trained and frozen TESSERA dual encoder (excluding the projector) is used. For each 10~meter pixel on the globe and for each year:
\begin{enumerate}
    \item The full Sentinel-1 and Sentinel-2 time series data at 10~meter resolution are acquired and pre-processed to form d-pixels. Unlike pre-training, no spatial downsampling is performed at this stage.
    \item A fixed number of 40 timesteps is sampled from the valid observations within the year for both Sentinel-1 and Sentinel-2 data, along with their DOY positional encodings.
    \item These sampled time series are fed into their respective frozen TESSERA encoders.
    \item The outputs from the S1 and S2 encoders are fused by the MLP, producing a 128-dimensional embedding vector for that pixel for that year.
\end{enumerate}
This process is repeated for all land pixels globally to create an annual embeddings map of shape (H, W, 128), where H and W are the dimensions of the global 10~meter grid.

\subsection{Spatial Context: Pixel-wise vs. Patch-based Approaches}
\label{sec:spatial_context}
In this section, we elaborate on the rationale behind TESSERA's pixel-wise design compared to traditional patch-based architectures. By maintaining pixel-wise independence, TESSERA avoids embedding fixed spatial priors, which preserves the integrity of raw spectral signatures and ensures downstream performance is driven by fundamental signal characteristics rather than pre-determined structural assumptions.

\begin{figure}[h]
    \centering
    \includegraphics[width=\linewidth]{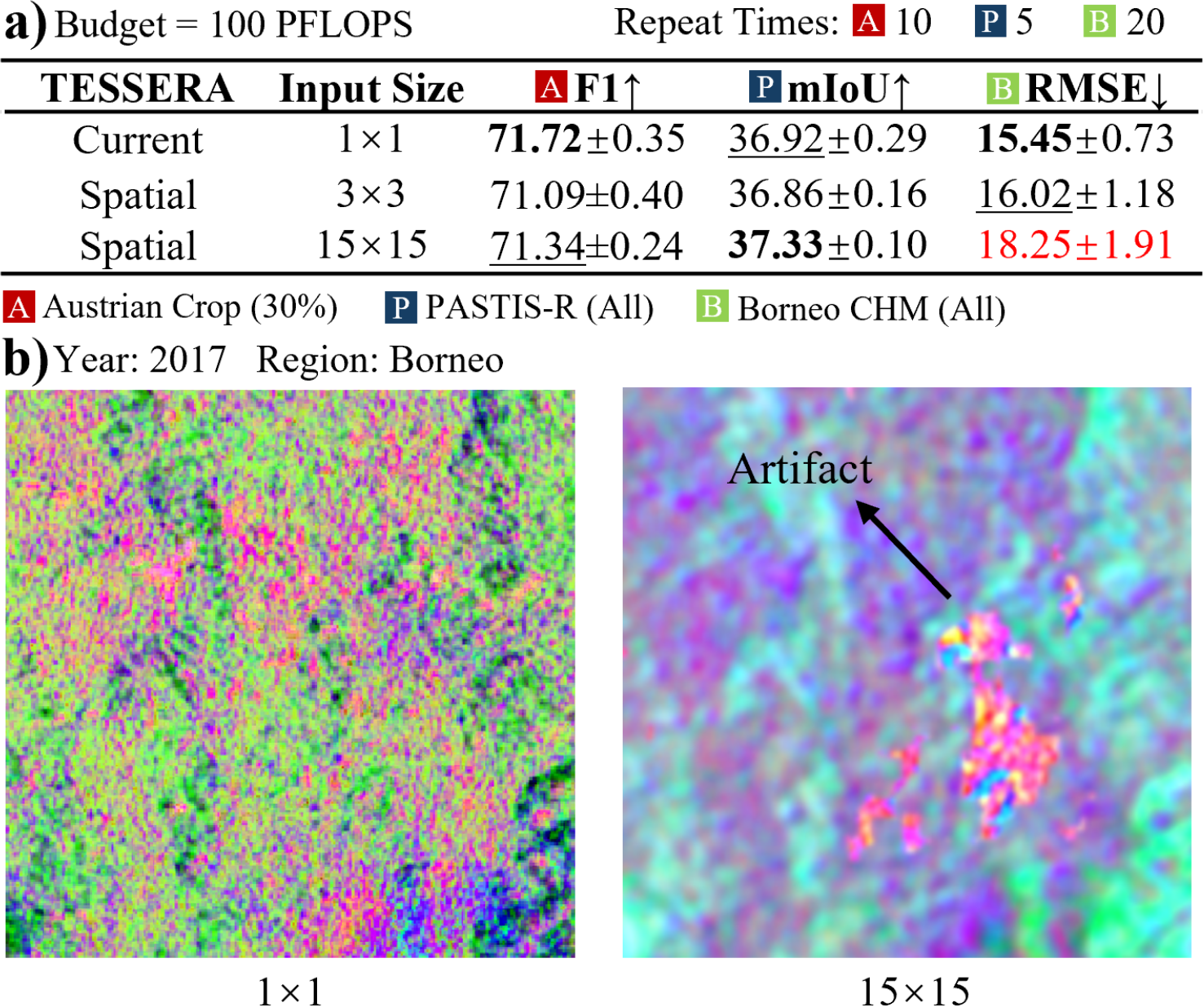} 
    \caption{\textbf{Spatial Context Ablation.} (a) Quantitative comparison detailing performance differences when introducing patch-based spatial priors. (b) Visualization of the generated embeddings in the Borneo region.}
    \label{fig:spatial_ablation}
\end{figure}

Introducing spatial context via fixed patches (Spatial variant) introduces significant vulnerabilities, particularly in regions with high cloud frequency such as Borneo. As the quantitative comparisons demonstrate (\Cref{fig:spatial_ablation}a), introducing patch-based spatial priors can negatively impact the robustness of the representations. Furthermore, filtering for strictly clear spatial patches results in an exponential decrease in data availability, leading to severe embedding artifacts. This vulnerability is clearly visible in the embedding visualizations of the Borneo region (\Cref{fig:spatial_ablation}b), where imposing a spatial patch requirement drastically reduces valid data and creates massive spatial voids. By extracting features at the pixel level, TESSERA allows researchers to incorporate domain-specific spatial constraints \textit{post hoc} without compromising the foundational embeddings.

\subsection{Geographic Generalization}
\label{sec:africa_generalization}
To demonstrate TESSERA's ability to generalize to diverse phenological regions and smallholder agricultural settings, we evaluate our model on the CropHarvest Togo dataset~\cite{tseng2021cropharvest} and the Ethiopia Crop classification dataset ~\cite{blasch2024ethiopian}. 
Despite using only a continuous DoY sinusoidal encoding as a weak temporal prior, TESSERA correctly handles local seasonality driven primarily by spectral trends.

\Cref{tab:africa_benchmark} compares TESSERA against other baseline foundation models. TESSERA significantly outperforms CROMA and AlphaEarth on both benchmarks, demonstrating superior generalization in challenging zero-shot and low-data transfer scenarios characteristic of Sub-Saharan Africa. 

\begin{table}[h]
\centering
\caption{\textbf{Downstream Performance on Sub-Saharan African Benchmarks.} Evaluation using F1 scores (\%) for the Togo (CropHarvest) and Accuary (\%) for the Ethiopia Crop dataset. TESSERA demonstrates superior adaptability to local seasonalities compared to existing models.}
\label{tab:africa_benchmark}
\begin{tabular}{lcc}
\toprule
\textbf{Model} & \textbf{Togo (F1 \%)} & \textbf{Ethiopia (Acc \%)} \\
\midrule
CROMA & 79.15 & / \\
AlphaEarth & / & $\sim$41 \\
\textbf{TESSERA (Ours)} & \textbf{83.01} & \textbf{44.92} \\
\bottomrule
\end{tabular}
\end{table}

\subsection{Scaling with data and network size}
To identify an optimal and efficient architecture for TESSERA, and to understand the scaling properties of our model, we conducted a series of experiments using the Austrian crop classification task as the downstream task. Our investigation focused on two main areas: the impact of the projector network's dimensions and the overall model scaling laws concerning performance versus computational cost (\Cref{fig:scaling_laws}).

Our experiments revealed that the performance of the final embeddings on the chosen downstream task of crop classification in Austria was relatively insensitive to the size of the Transformer encoders. We found that using 4 transformer layers, each with 4 attention heads, provided a strong baseline without incurring excessive computational overhead during inference. This makes the feature-extracting part of the model computationally efficient.

In contrast, the architecture of the projector network proved to be critical for effective self-supervised learning. As detailed in Section~\ref{sec:projector}, the full projector is a six-layer MLP, consisting of an input and an output layer for accommodating dimensional changes, and four core hidden layers. Our parameter sweep focused on this computationally intensive hidden block. We varied the number of these hidden layers (referred to as 'depth' in ~\Cref{fig:scaling_laws} a) and their width (number of neurons per layer), with results shown in ~\Cref{fig:scaling_laws} a. As discussed by Zbontar et al.~\cite{Zbontar2021BarlowTS}, deeper and wider projectors consistently improved performance, as measured by the validation F1 score in the downstream task. The highest validation score was achieved with the largest configuration tested (8 hidden layers, 32,768 neurons wide). However, this configuration led to an out-of-memory error with our standard global batch size; therefore, we halved the batch size to 16,384 to enable stable training for this specific setup, indicated by the dagger symbol (\textdagger) in the figure. For the final TESSERA v1.0 model, we selected the configuration with four hidden layers and a width of 16,384, marked with a star (*), as it provided a strong balance between high performance and computational efficiency.

Furthermore, we analysed the relationship between model performance, the number of trainable parameters, and the total training computation (FLOPs), as shown in ~\Cref{fig:scaling_laws} b. The results demonstrate a clear scaling trend: model performance improves predictably with increases in model size and computational budget. This confirms that TESSERA's performance can be further enhanced with greater computational resources, following established scaling laws for large neural networks. The final TESSERA v1.0 model is marked with a star (*) on this curve, highlighting its position as a model that balances state-of-the-art performance with cost-effective training requirements.

\subsubsection{Ablation study details}


To validate the contribution of each key component within the TESSERA framework, we performed an ablation study. We systematically removed or disabled individual components from our full model (the baseline) and evaluated the impact on the 100x downsampled Austrian crop classification dataset.

Performance was evaluated using two metrics: the standard validation overall F1 score, and a measure of the effective rank of the learned embeddings, which we refer to as the RankMe score \cite{garrido2023rankme}. The RankMe score quantifies the richness of the embeddings by measuring how uniformly the information is distributed across the feature dimensions. Given a batch of embeddings represented by the matrix $Z \in \mathbb{R}^{N \times D}$, where $N$ is the batch size and $D$ is the embedding dimension, we first compute its singular values $s_1, s_2, \ldots, s_D$. These are normalized to form a probability distribution $p_i = s_i / \sum_{j} s_j$. The RankMe score is the Shannon entropy of this distribution, normalized by the maximum possible entropy:
\begin{equation}
\mathcal{R}_{\text{RankMe}} = \frac{-\sum_{i=1}^{D} p_i \log(p_i)}{\log(D)}
\label{eq:rankme}
\end{equation}
A score closer to 1 indicates that the embeddings are of higher effective rank, utilizing the full dimensionality of the feature space more effectively to create richer, less redundant representations.

Our analysis reveals several key insights into the model's architecture and training strategy. The most critical component turns out to be the global data shuffling pipeline, which is designed to break spatial autocorrelation and present the model with geographically diverse mini-batches. Removing mixup regularization also severely degrades results. When both are removed, the model's performance collapses, showing that the combination of global shuffling and mixup regularization is needed to prevent overfitting and to learn generalizable robust representations.

With removal of Sentinel-1 data, task performance decreases moderately, demonstrating some gain from the fusion of optical and SAR data yields. However, performance is still relatively good even with only Sentinel-2 data. This highlights both TESSERA's overall robustness and its continued efficacy in regions or historical periods where Sentinel-1 data are sparse or unavailable, which is not uncommon in practice.

Finally, the effect of quantization is particularly noteworthy. We employ Quantization-Aware Training (QAT) to convert the embeddings from 32-bit floating-point (float32) to 8-bit integer (int8) format. Removing this quantization step (i.e. using full float32 precision) results in a marginal performance improvement. This demonstrates the success of our QAT strategy: we achieve a nearly four-fold reduction in the storage and bandwidth requirements for our final data products, with a negligible impact on embedding quality. This underlies our "Embeddings-as-Data" approach, making global-scale high-resolution analysis practical: Without this, each year's embeddings would consume ~1PB, making the hosting of 8 years of embedding costly and likely unachievable.

\subsubsection{Justification for the Number of Sampled Timesteps ($L$)}
\label{sec:supp_justification_L}

\begin{table}[t]
\centering
\caption{Parameter sweep of the number of sampled timesteps ($L$). We evaluate the trade-off between F1-score on the Austrian crop classification dataset and inference speed.}
\label{tab:ablation_L_timesteps}
\vspace{-2mm}
\scriptsize
\setlength{\tabcolsep}{4pt} 
\renewcommand{\arraystretch}{0.8}
\begin{tabular}{c c c c c}
\toprule
\textbf{$L$} & \textbf{\begin{tabular}[c]{@{}c@{}}Inference Speed \\ (samples/sec)\end{tabular}} & \textbf{Encoder Param.} & \textbf{Training Samples} & \textbf{F1 Score (\%)} \\
\midrule
15 & 1509 & \multirow{6}{*}{$\sim$36.4M} & \multirow{6}{*}{22M} & 65.04 \\
20 & 1452 & & & 66.58 \\
30 & 1276 & & & 71.43 \\
\textbf{40} & \textbf{1068} & & & \textbf{73.34} \\
50 & 724 & & & 73.69 \\
60 & 497 & & & 73.77 \\
\bottomrule
\end{tabular}
\vspace{-2mm}
\end{table}

\begin{figure}[t]
    \centering
    \includegraphics[width=1\linewidth]{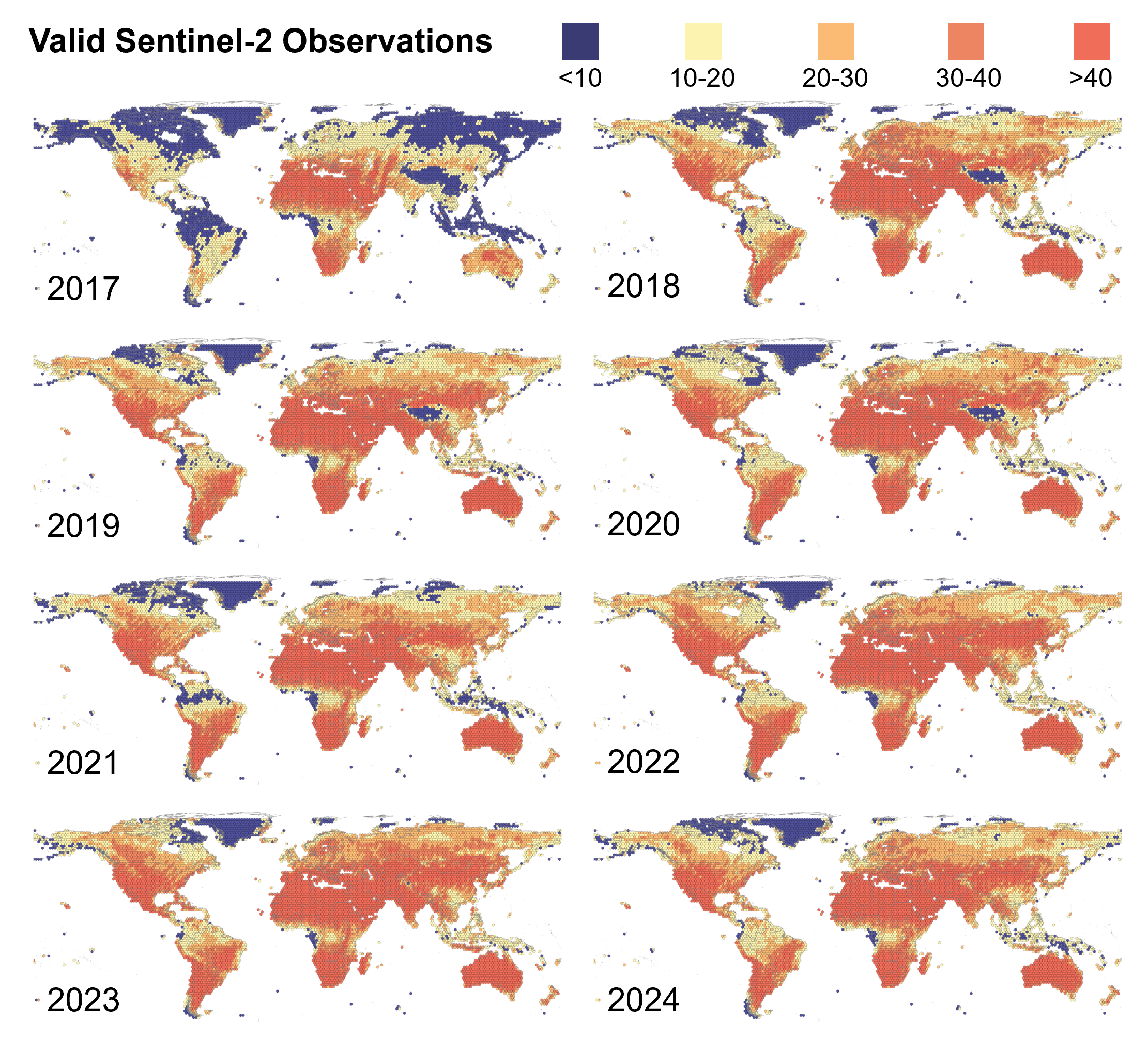}
    \caption{A distribution map of the number of valid Sentinel-2 observations from 2017 to 2024. Valid observations are defined as those that are free of clouds, water bodies, and snow.}
    \label{fig:valid_s2_distribution}
\end{figure}

As detailed in the main paper, TESSERA is trained using sparse random temporal sampling to enforce invariance to the selection of valid observations. A critical hyperparameter in this process is $L$, the number of valid timesteps sampled. We finalized our choice at $L=40$. This decision is supported by two primary justifications:

\begin{enumerate}
    \item \textbf{Performance vs. Efficiency Trade-off:} We conducted a comprehensive parameter sweep on the Austrian crop classification dataset to determine the optimal balance between model performance and computational cost (i.e., inference speed). As demonstrated in ~\Cref{tab:ablation_L_timesteps}, setting $L=40$ achieves the best trade-off. While further increasing $L$ (e.g., to 50 or 60) results in only marginal or negligible performance gains, it increases the computational overhead during inference.
    
    \item \textbf{Global Data Representativeness:} Our choice is also empirically grounded in global-scale data statistics. We analyzed the global distribution of valid Sentinel-2 observations acquired between 2017 and 2024 and found that the global average number of valid observations per pixel is approximately 37.6. This
    confirms that $L=40$ is not only efficient but also a robust and representative value for global-scale modeling, closely aligning with the average valid Sentinel-2 data availability worldwide (see \Cref{fig:valid_s2_distribution}).
\end{enumerate}

Furthermore, we observed that downstream task performance is largely insensitive to the number of Sentinel-1 observations sampled, likely due to the higher temporal density and all-weather capabilities of SAR. Therefore, for the sake of architectural consistency and to simplify the multi-modal training pipeline, we also set the number of sampled timesteps for Sentinel-1 to $L=40$.

\subsection{Downstream Task Application Methodology}
\label{app:downstream_methodology}

\subsubsection{TESSERA}
A core motivation for self-supervised learning with foundation models is the creation of task-agnostic feature embeddings that can be effectively transferred to various downstream tasks, particularly in scenarios with limited labelled data. Having pre-trained TESSERA on large unlabelled datasets, we evaluated the utility of its learned embeddings across different remote sensing applications. The evaluation methodology involves using pre-trained TESSERA encoders as fixed feature extractors.

\begin{enumerate}
    \item \textbf{Download Embeddings for the Region of Interest}: The \textsc{GeoTessera} Python library allows users to download embeddings for a
    desired region and year in the form of a numpy array.
    \item \textbf{Prepare labelled Downstream Data}: The labelled dataset for the target task (e.g., pixel-level crop-type labels, canopy height measurements, or land use change polygons) is prepared.
    \item \textbf{Design Task-Specific Head}: A lightweight, task-specific neural network module (the "head") is designed. This head takes the extracted TESSERA embeddings as input.
    \begin{itemize}
        \item For pixel-wise classification (e.g., crop classification), the head is typically a shallow MLP (1-3 layers) ending in a softmax output layer.
        \item For pixel-wise regression (e.g., canopy height regression), the head is usually an MLP ending in a single linear output neuron.
        \item For tasks requiring spatial context from the embeddings (e.g., canopy height mapping over an area, semantic segmentation), the input to the head can be a patch of TESSERA embeddings (e.g., $64 \times 64 \times 128$). The head might then be a convolutional architecture, such as a UNet, that processes these spatial feature maps to produce dense predictions.
    \end{itemize}
    \item \textbf{Train Downstream Head}: Only the parameters of this newly defined task head are trained using the extracted TESSERA embeddings as input features and the corresponding labels. Standard supervised learning techniques, optimizers (e.g., Adam), and task-appropriate loss functions (e.g., Cross-Entropy for classification, Mean Squared Error for regression) are used. This training typically requires significantly less labelled data and computational power compared to training a deep model from scratch.
    \item \textbf{Evaluation}: Once the head is trained, inference is performed on a test set by extracting TESSERA embeddings for the test samples and passing them through the trained head. Performance is evaluated using standard metrics relevant to the task.
\end{enumerate}
This workflow allows the use of TESSERA embeddings in a range of diverse applications, demonstrating its role as a foundational model for geospatial analysis.

TESSERA users who would like to generate their own embeddings can load the frozen
weights from the saved pre-training checkpoint into the TESSERA dual encoder architecture. For each input sample (e.g., pixel, object, or patch) in the region of interest, its corresponding Sentinel-1 and Sentinel-2 time series data for the relevant year must be downloaded and processed using the d-pixel creation pipeline. These d-pixels are passed through the frozen encoders and their outputs are fused by the MLP to generate the final 128-dimensional TESSERA embedding for that sample. To enhance stability, multiple embeddings can be created with different random temporal samples and averaged.

\subsubsection{Presto \& AlphaEarth}
\begin{figure}[t]
    \centering
    \includegraphics[width=1\linewidth]{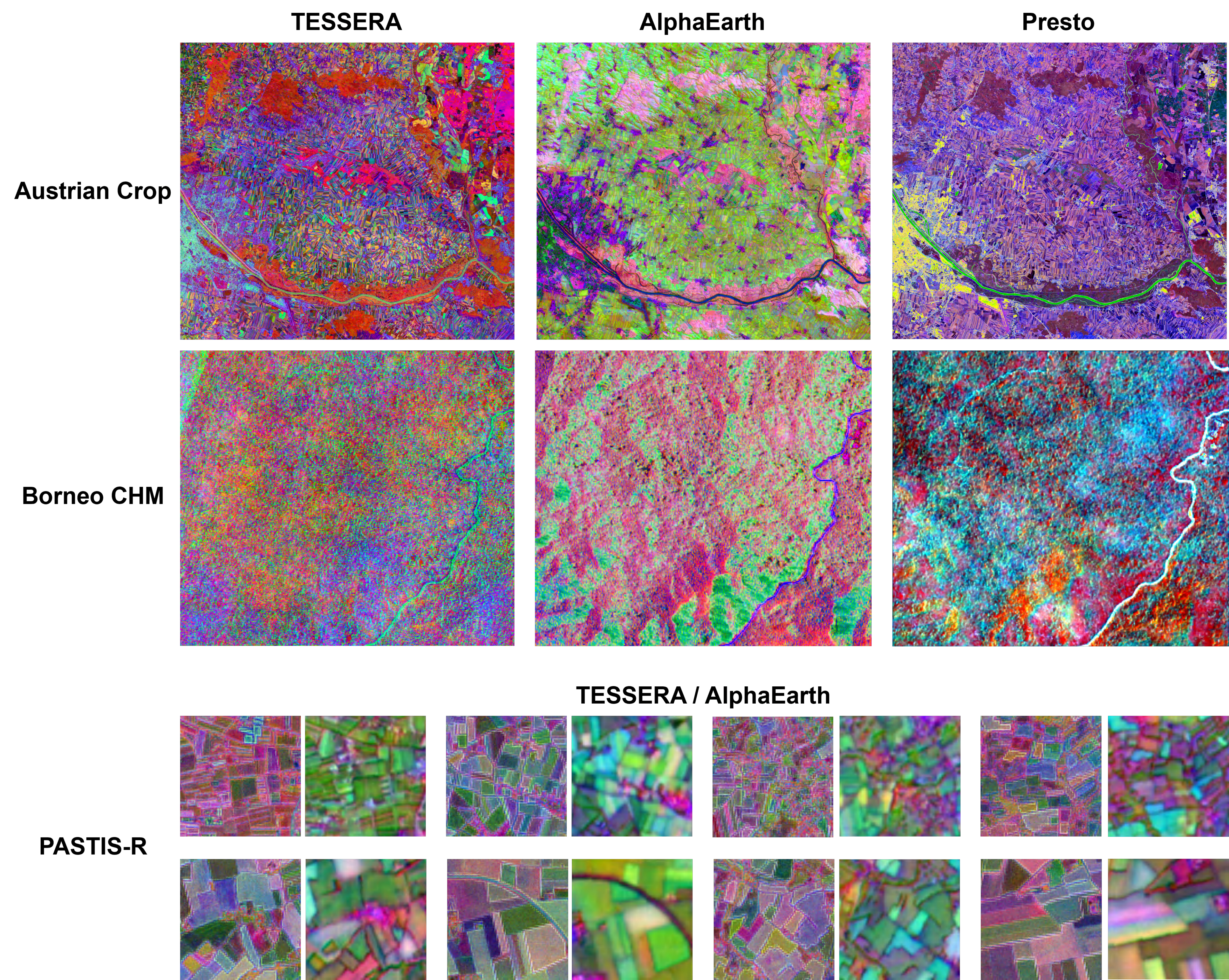}
    \caption{The visualization of embeddings for some downstream tasks of TESSERA, AlphaEarth, and Presto}
    \label{fig:emb_vis}
\end{figure}

For Presto and AlphaEarth, we adopted a methodology conceptually similar to TESSERA's, focusing on evaluating their publicly available embeddings.
For Presto, we used the official Google Earth Engine (GEE) scripts provided by the authors to generate embeddings for the regions of interest. The data preprocessing was executed on GEE, and inference was subsequently performed on Google Vertex AI, loading the officially released pre-trained model weights.
For AlphaEarth, we directly used the publicly available Google AlphaEarth Satellite Embedding V1 dataset~\cite{SatelliteEmbeddingV1}.
In both cases, the pre-computed embeddings served as the input features. The visualization of embeddings for some downstream tasks of TESSERA, AlphaEarth, and Presto are shown in \Cref{fig:emb_vis}.
We then designed and trained lightweight task-specific heads (e.g., MLPs) for each downstream task, following the same principle as described for TESSERA.

\subsubsection{Other RSFMs}
To evaluate the performance of other RSFMs, we used the standardized \textit{Pangaea} evaluation framework. This framework provides a consistent protocol for benchmarking RSFMs across a suite of diverse downstream tasks. We evaluated these models on the Austrian Crop (Segmentation), PASTIS-R, Biomassters, and Borneo CHM datasets.
Unlike our embedding-focused approach for TESSERA, the Pangaea protocol typically involves training task-specific decoders on top of the pre-trained encoder features. For classification tasks, a simple \texttt{LinearClassifier} head was attached and trained. For the regression and segmentation tasks, heavier decoders such as \texttt{UperNet+LTAE} or a baseline \texttt{Unet} were employed, as specified in the benchmark's guidelines. The exact downstream task heads used for these models are detailed in ~\Cref{tab:model_heads}.

\begin{table}[t]
\centering
\caption{\textbf{Downstream task head specification.} This table lists the decoders (heads) used for different tasks across all models. Note that for each model listed, the Regression and Segmentation tasks use the same head.}
\label{tab:model_heads}
\vspace{-2mm}
\scriptsize
\setlength{\tabcolsep}{4pt} 
\renewcommand{\arraystretch}{0.8}
\begin{tabular}{l c c}
\toprule
\textbf{Model} & \textbf{Classification Head} & \textbf{Regression \& Segmentation Head} \\
\midrule
CROMA & LinearClassifier & UperNet+LTAE \\
DOFA & LinearClassifier & UperNet+LTAE \\
Prithvi & LinearClassifier & UperNet+LTAE \\
RemoteCLIP & LinearClassifier & UperNet+LTAE \\
SatlasNet & LinearClassifier & UperNet+LTAE \\
Scale-MAE & LinearClassifier & UperNet+LTAE \\
SpectralGPT & LinearClassifier & UperNet+LTAE \\
Galileo & LinearClassifier & UperNet+LTAE \\
Skysense & LinearClassifier & UperNet+LTAE \\
ViT Baseline & LinearClassifier & UperNet+LTAE \\
UNet Baseline & LinearClassifier & Unet \\
\midrule
Presto & LinearClassifier & Unet \\
AlphaEarth & LinearClassifier & Unet \\
TESSERA & LinearClassifier & Unet \\
\bottomrule
\end{tabular}
\vspace{-2mm}
\end{table}

\subsection{Downstream Task: Austrian Crop Classification}
\label{sec:detail:austriancrop}

\paragraph{Dataset and Preprocessing}
We used the publicly available INVEKOS dataset for Austria, focusing on the 2022 growing season \cite{noauthor_invekos_nodate}. The data set originally contained 154 crop types, which we grouped into 17 broader classes (for example, combining sugar beet and feed beet) based on phenological similarity and sample availability to ensure robust training and evaluation.

When constructing the pixel-level classification dataset, we use fields as the unit for data splitting. Each field is assigned a unique identifier along with its corresponding area. We allocate pixels from fields constituting X\% of the total area to the training set, while the remaining (1-X)\% of the total field area is divided such that 1/7 is used for validation and the rest for testing. For training, validation, and test splits, we ensure that all 17 crop types are represented in each split.

For the patch-based dataset designed for segmentation tasks, we first divide the region of interest vertically into five equal parts. From the topmost section, we extract 1000 patches of 640m × 640m as the test set. An equal number of patches with the same dimensions are cropped from the section immediately below for the validation set. From the remaining three sections, 3000 patches of the same size are sampled to form the training set. All patches maintain a spatial resolution of 10 meters.

\paragraph{Pixel-wise Classification Baselines}
To provide a comprehensive performance comparison, we implemented three distinct models for pixel-wise classification:
\begin{itemize}
    \item \textbf{TESSERA + MLP}: The primary model, where the frozen 128-dimensional TESSERA embeddings were used as input to a simple MLP. The MLP consisted of two hidden layers with 256 and 128 neurons, respectively, using ReLU activation functions, followed by a softmax output layer for the 17 classes.
    \item \textbf{Presto + MLP}: For a direct and fair comparison with a leading foundation model, we used the official pre-trained Presto model~\cite{tseng_lightweight_2024} to generate its pixel embeddings. These embeddings were then fed into an MLP head identical to the one used for TESSERA.
    \item \textbf{AlphaEarth + MLP}: To include another key foundation model, we used the publicly available Google AphaEarth Satellite Embedding V1 dataset~\cite{SatelliteEmbeddingV1}. These embeddings were similarly fed into an MLP head identical to the one used for TESSERA.
    \item \textbf{Random Forest}: As a traditional baseline, we trained a Random Forest (RF) classifier directly on raw time-series data. For each pixel, we used all available Sentinel-1 (2 polarizations) and Sentinel-2 (10 spectral bands) observations throughout the year. The temporal and spectral/polarization dimensions were flattened and concatenated to form a single 1256-dimensional feature vector. The RF model was configured with 100 trees, while all other hyperparameters retained their default values as implemented in the \textsc{scikit-learn} library.
\end{itemize}
For the experiments shown in the main text's figure, we trained these models on randomly selected subsets of the data (from 1\% to 30\% for panel a; specified samples-per-class for panel b), using a fixed validation set for hyperparameter tuning and a held-out test set for final evaluation.

\begin{table*}[t]
\centering
\caption{\textbf{Pre-training hyperparameters for the TESSERA v1.0 model.}}
\label{tab:hyperparameters}
\vspace{-2mm}
\setlength{\tabcolsep}{6pt}
\renewcommand{\arraystretch}{0.8}
\begin{tabular}{lcl}
\toprule
\textbf{Description} & \textbf{Symbol} &
\textbf{Value} \\
\midrule
\multicolumn{3}{c}{\textbf{Optimizer \& Schedule}} \\
Optimizer & - & AdamW \\
Batch Size (Per-GPU) & $B_{local}$ & 2,048 \\
Batch Size (Global) & $B_{global}$ & 32,768 (16 GPUs $\times$ 2,048) \\
Epochs & - & 1 \\
Base Learning Rate & $\eta$ & 0.002 \\
Weight Decay & $\lambda_{wd}$ & $1 \times 10^{-6}$
\\
Learning Rate Schedule & - & Linear warmup (10\%)
followed by cosine decay \\
Gradient Clipping Norm & - & 2.0 \\
\midrule
\multicolumn{3}{c}{\textbf{Loss Function}} \\
Barlow Twins Lambda & $\lambda_{BT}$ & $5 \times
10^{-3}$ \\
Mixup Lambda & $\lambda_{mix}$ & 1.0 \\
\midrule
\multicolumn{3}{c}{\textbf{Architecture \& Data}} \\
Sequence Length & $L$ & 40 \\
Representation Dimension & $D_{repr}$ & 128 \\
Projector Output Dimension & $D_{proj}$ & 16,384 \\
Quantization Bits (QAT) & - & 8 \\
\bottomrule
\end{tabular}
\vspace{-2mm}
\end{table*}

\paragraph{Patch-wise Semantic Segmentation}
To assess spatial-contextual performance, we conducted a semantic segmentation experiment. The approach varied based on the foundation model's architecture:
\begin{itemize}
    \item For the foundation models capable of generating pixel-wise embeddings, namely \textbf{TESSERA}, \textbf{Presto}~\cite{tseng_lightweight_2024}, and \textbf{AlphaEarth}~\cite{SatelliteEmbeddingV1}, a distinct workflow was used. We first construct patches from their pixel-level embeddings (e.g., a data cube of size $H \times W \times C$). These patches were then fed into a standard \textbf{UNet} architecture~\cite{ronnebergerUNetConvolutionalNetworks2015}, which learns to model the spatial relationships between the embeddings to produce a segmentation map. It is noteworthy that while TESSERA and Presto's base embeddings do not explicitly model spatial context, AlphaEarth's do. However, their shared ability to generate pixel-level outputs makes them suitable for this UNet-based downstream approach.
    \item For other foundation models that are inherently patch-based (e.g., Prithvi \cite{szwarcman2025prithvieo20versatilemultitemporalfoundation}, Satlas \cite{Bastani2023}), their encoders already process image patches. Therefore, for these models, we only attach and train a \textbf{UPerNet decoder head} to their encoders to generate the final segmentation output. We did not freeze the encoders during the fine-tuning.
\end{itemize}
Performance was measured using mean Intersection over Union (mIoU) and macro F1 scores. A visual comparison for example patches is provided in Supplementary ~\Cref{fig:sup_austrian_patch}.

\begin{figure}[!h]
    \centering
    \includegraphics[width=\linewidth]{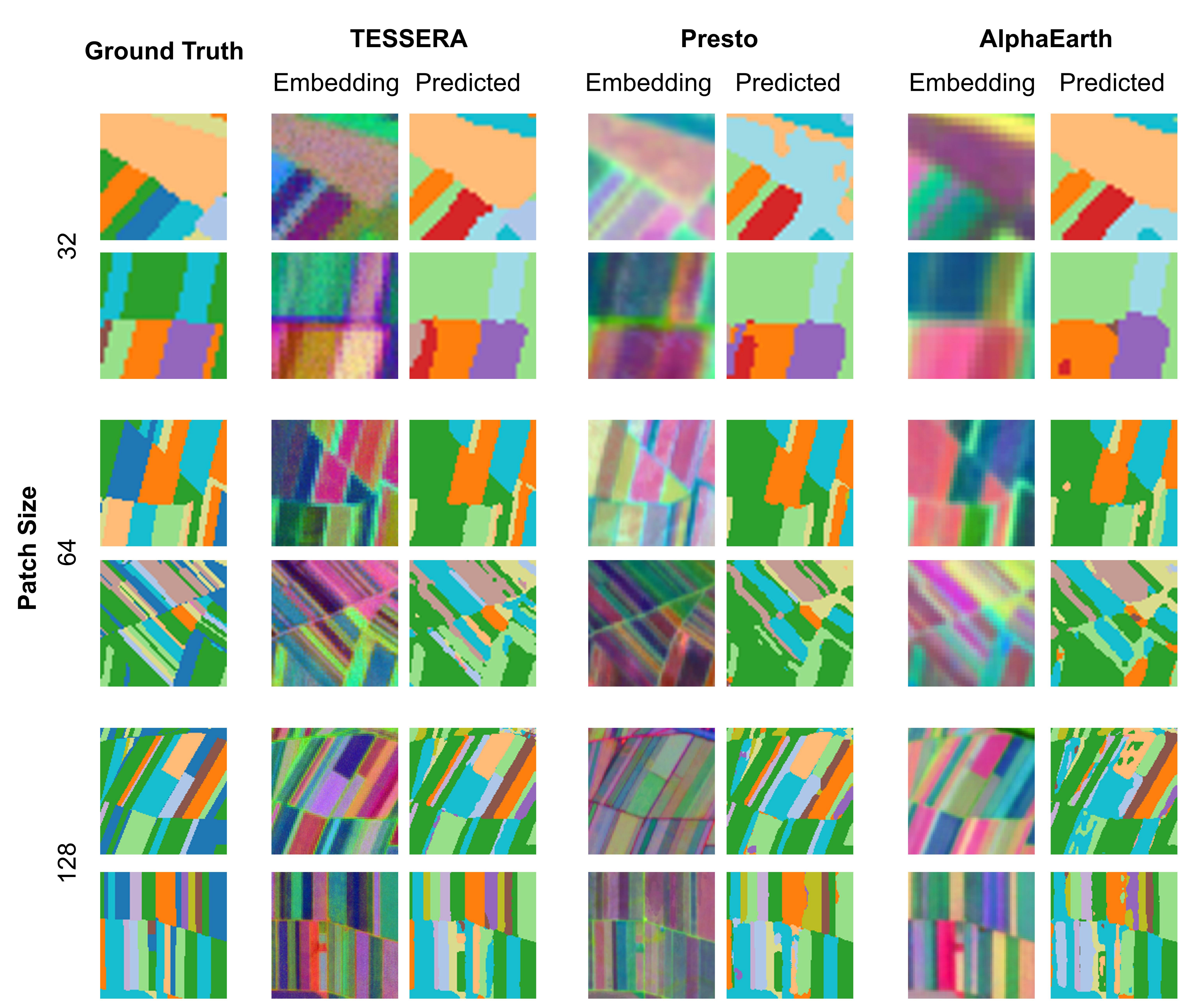}
    \caption{\textbf{TESSERA predictions most closely resemble the ground truth for patch-based semantic segmentation.} Each row corresponds to a different patch size (32x32, 64x64, 128x128). The first column shows the ground truth segmentation. For each model (TESSERA, Presto, and AlphaEarth), two columns are displayed: a PCA visualization of the input embeddings for the patch (labelled 'Embedding') and the final predicted segmentation map (labelled 'Predicted').}
    \label{fig:sup_austrian_patch}
\end{figure}

\paragraph{Embedding Space Analysis}

\begin{figure}[!h]
    \centering
    \includegraphics[width=1\linewidth]{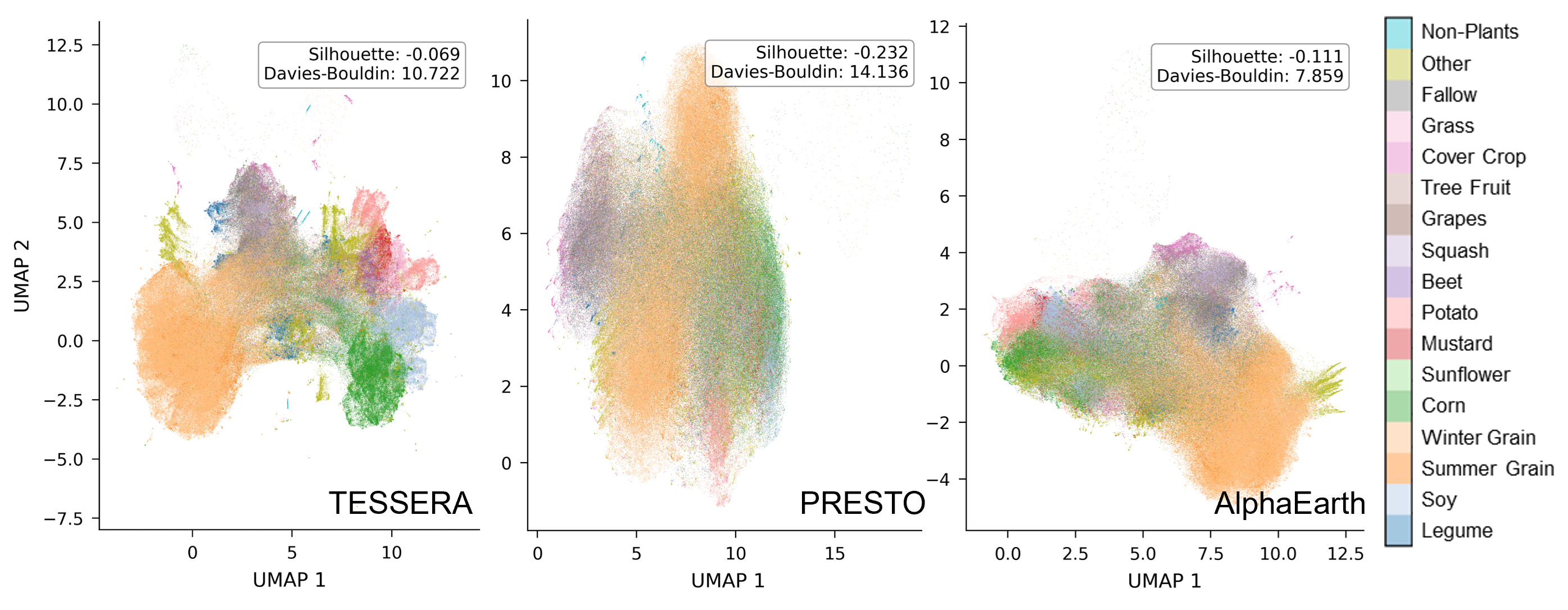}
    \caption{UMAP visualization of the embedding spaces for TESSERA, Presto, and AlphaEarth for 17 crop classes. TESSERA's embeddings exhibit clearer separation and more coherent clustering visually.}
    \label{fig:umap}
\end{figure}

The 2D visualizations shown in \Cref{fig:umap} were generated by applying the Uniform Manifold Approximation and Projection (UMAP) \cite{mcinnes2018umap}  algorithm to the embeddings of every pixel within the Austrian study area. 
UMAP is a non-linear dimensionality reduction technique based on Riemannian geometry and algebraic topology. It constructs a weighted k-nearest neighbour graph in the original high-dimensional space, then optimizes a low-dimensional projection to primarily preserve local distances and local structure. UMAP, like other non-linear techniques such as t-SNE and LargeVis, is designed to preserve local relationships within high-dimensional data; however, it is also able to retain more of the global structure, making it particularly effective for understanding both fine-scale and broader patterns in the embedding space.
In this work, the entire embedding map (for example, a shape array $H \times W \times 128$, where H and W are the height and width of the region) was used as a direct input to UMAP for the TESSERA, Presto, and AlphaEarth embeddings.

To quantitatively measure the quality of the clustering in the original 128-dimensional space, we calculated the Silhouette score and the Davies-Bouldin Index.
\begin{itemize}
    \item The \textbf{Silhouette score}, $s(i)$, for a single data point $i$ measures how similar it is to its own cluster compared to other clusters. It is defined as:
    \begin{equation}
        s(i) = \frac{b(i) - a(i)}{\max(a(i), b(i))}
    \end{equation}
    where $a(i)$ is the mean distance between $i$ and all other points in the same cluster, and $b(i)$ is the mean distance from $i$ to all points in the nearest neighbouring cluster. The score ranges from -1 to 1, where a high value indicates that the object is well matched to its own cluster and poorly matched to neighboring clusters.
    \item The \textbf{Davies-Bouldin Index} (DBI) evaluates clustering quality by computing the ratio of within-cluster scatter to between-cluster separation. For a set of $k$ clusters, it is defined as:
    \begin{equation}
        \text{DBI} = \frac{1}{k} \sum_{i=1}^{k} \max_{j \neq i} \left( \frac{\sigma_i + \sigma_j}{d(c_i, c_j)} \right)
    \end{equation}
    where $\sigma_i$ is the average distance of all points in cluster $i$ to their centroid $c_i$, and $d(c_i, c_j)$ is the distance between the centroids of clusters $i$ and $j$. Lower DBI values indicate better clustering, with a score of 0 representing the ideal case where the clusters are compact and well separated.
\end{itemize}

\paragraph{Impact of Seasonal Windows}

We analyze the impact of using different temporal windows on the model's performance for the downstream task of crop classification on the Austrian dataset. As shown in \Cref{tab:seasonal_ablation}, we conduct an ablation by systematically excluding one of the four meteorological seasons (Spring, Summer, Autumn, Winter) from the input time series. The results reveal that, for this specific regional dataset, excluding time steps from the winter season improves the classification F1 score from 78.94\% (all seasons included) to $\mathbf{79.83\%}$. This suggests that the dormant winter period, which is typically characterized by low vegetation phenological activity and is prone to cloud cover, snow, or reduced solar zenith angle effects, introduces a substantial amount of non-discriminative noise. By excluding the winter data, the model can focus its attention and resources on the most informative growing seasons, which allows for the learning of a more concise and effective temporal representation for crop type identification.

\begin{table}[t]
    \centering
    \caption{\textbf{Ablation on the Austrian Crop dataset: Seasonal Data Inclusion.} We report the \textbf{Validation F1} score (higher $\uparrow$ is better) for the crop classification task. The columns for Spring to Winter use \yes ~to indicate inclusion and \no  ~to indicate exclusion. Parentheses show change vs.~the Full Model.}
    \label{tab:seasonal_ablation}
    \vspace{-2mm}
    \setlength{\tabcolsep}{4.5pt}
    \renewcommand{\arraystretch}{0.9}
    \resizebox{\columnwidth}{!}{%
    \begin{tabular}{cccccc}
        \toprule
        \textbf{Configuration} & \textbf{Spring} & \textbf{Summer} & \textbf{Autumn} & \textbf{Winter} & \textbf{F1 ($\uparrow$)} \\
        \midrule
        \textbf{Full Model (Baseline)} & \yes & \yes & \yes & \yes & 78.94 \\
        \midrule
        w/o Spring & \no & \yes & \yes & \yes & 78.62 (\textcolor{red}{-0.32}) \\
        w/o Summer & \yes & \no & \yes & \yes & 78.10 (\textcolor{red}{-0.84}) \\
        w/o Autumn & \yes & \yes & \no & \yes & 78.22 (\textcolor{red}{-0.72}) \\
        \textbf{w/o Winter} & \yes & \yes & \yes & \no & \textbf{79.83} (\textcolor{green}{+0.89}) \\
        \bottomrule
    \end{tabular}%
    }
    \vspace{-2mm}
\end{table}

\subsection{Downstream Task: Canopy height estimation}
\paragraph{Study Areas and Ground Truth Datasets}
Canopy height prediction was evaluated in a 5$\times$6~km area in the Danum Valley, Borneo. This region is covered in tropical rainforest dominated by species in the Dipterocarpaceae family, which hosts some of the tallest trees in the tropics \cite{shenkinWorldsTallestTropical2019, piponiotDistributionBiomassDynamics2022}. Ground truth data was derived from airborne laser scanning (ALS) performed in February 2020 \cite{coomesAirborneLiDARRGB2022}. ALS data provides detailed 3D point clouds of vegetation and terrain structure, which were processed using the ~\textsc{lspikefree} algorithm \cite{fischerRobustCharacterisationForest2024} to estimate top-of-canopy height at 1~m resolution. These canopy height rasters were then downsampled using mean pooling to 10~m to match the resolution of the TESSERA representations.

\paragraph{Modelling methods}
To predict the canopy height from the different embeddings, we used a standard U-Net architecture \cite{ronnebergerUNetConvolutionalNetworks2015}. It takes as input a 32 $\times$ 32 pixel patch (covering a 320 $\times$ 320 m area) of representations. For 128-dimensional embeddings (as in TESSERA and Presto), this results in a 32 $\times$ 32 $\times$ 128 data cube. The network outputs a single-band 32 $\times$ 32 raster of predicted canopy height values. The proposed U-Net contains approximately 30 million parameters and was trained for 200 epochs using a batch size of four patches. At the end of each epoch, the model was evaluated using validation data and the model with the best performance on the validation set was retained for the final evaluation. In practice, training usually converged within the first 20 epochs, with rare boosts in performance later in training. Model training was done in Python using PyTorch.

To ensure spatial generalization and avoid overfitting local patterns, the reference data were divided into training, validation, and test sets that are spatially disjoint. Each study region was divided into four spatial folds based on cardinal directions (bottom, top, left, right), each fold covering 50\% of the area. For each fold, the remaining half of the region was used for training and validation. To account for inherent model randomness, we trained three independent models per fold, resulting in 12 trained models for each foundation model. This strategy captures variability due to both spatial heterogeneity and training randomness. The final performance of the model was evaluated on the test data held using the following metrics: coefficient of determination (R²), root mean square error (RMSE), and mean bias.

\paragraph{Baselines}
TESSERA, Presto~\cite{tseng_lightweight_2024}, and AlphaEarth~\cite{SatelliteEmbeddingV1} embeddings were compared by training identical models using each representation as input, with all other settings held constant. Each model was trained for 200 epochs on the same data split, and the version achieving the best validation performance was selected for final evaluation on the held-out test set.

To benchmark performance against classical remote sensing approaches applied at the global scale, we compared the predictions against a global canopy height product: Tolan et al. [2024]\cite{tolanVeryHighResolution2024} (Meta). These maps were downloaded at 10 m resolution from Google Earth Engine and compared directly to the canopy height derived from ALS. Although there are some temporal mismatches between ALS acquisition (2020) and global products, we assume relative stability of canopy height in the short term in this region, which is reasonable given that the area was not logged or burned.

\begin{figure*}[t]
    \centering
    \includegraphics[width=\linewidth]{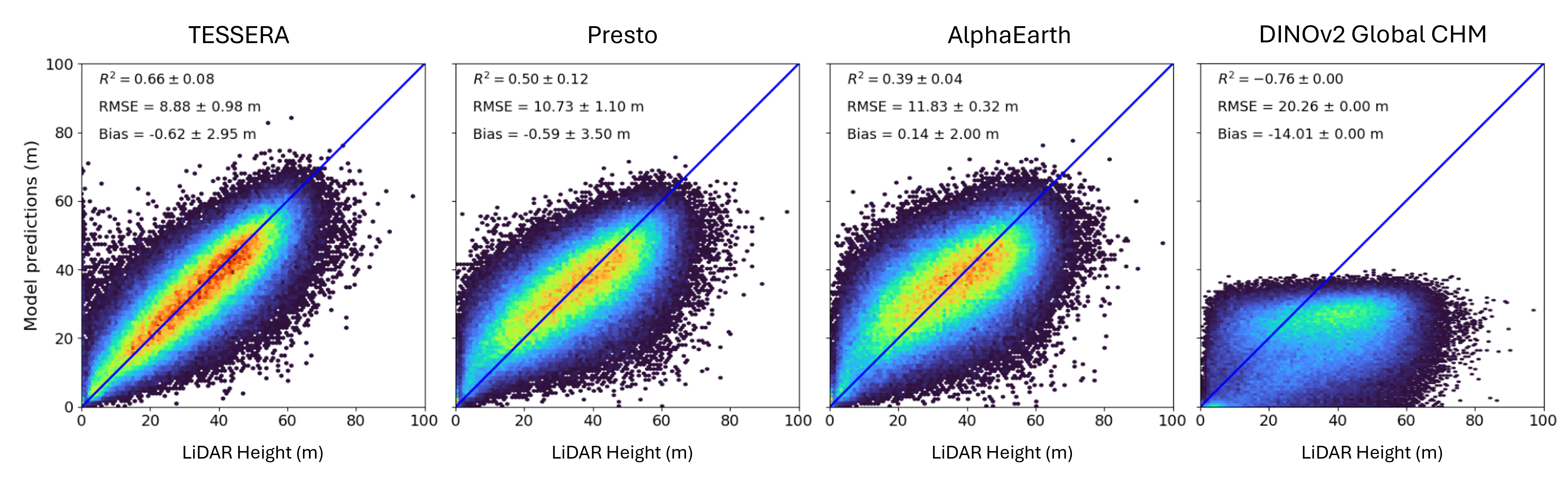}
    \caption{\textbf{Comparison of canopy height estimation performance across four models.}
    Predicted canopy height (y-axis) is compared against airborne LiDAR reference measurements (x-axis) in Danum Valley, Borneo. 
    The figure shows scatter-density plots for four model variants trained using different input representations: 
    \textbf{TESSERA}, \textbf{Presto}, \textbf{AlphaEarth}, and the \textbf{DINOv2 Global CHM} product.
    All models were evaluated using the same LiDAR reference dataset for consistency. 
    TESSERA achieves the highest coefficient of determination and lowest RMSE, demonstrating superior performance in structurally complex tropical canopies. 
    In contrast, the DINOv2 Global CHM product exhibits strong underestimation, large negative bias, and a near-zero or negative correlation, reflecting substantial saturation at lower canopy heights.
    A 1:1 reference line (blue) is included in each subplot for comparison.}
    \label{fig:canopy_height_short}
\end{figure*}

Finally, we trained the U-Net model using Sentinel-2 input as a baseline. This approach retained the same model architecture, patch size, and evaluation setup, but replaced embeddings with cloud-free annual median composites of Sentinel-2 imagery. Cloud masking and compositing were performed in GEE, resulting in 12 channel input images, 
with one channel for each Sentinel-2 spectral band. Although different Sentinel-2 bands have different spatial resolutions, all were resampled to 10~m to match the FM pipelines.

\subsection{Downstream Task: Above-ground biomass estimation}
\label{sec:detail:biomass}

\paragraph{Dataset}

For the AGB regression task, we used the ground truth data from 
the BioMassters dataset~\cite{Nascetti}. 
This dataset covers nearly 11,500 forest patches in Finland, with each patch representing a 2,560$\times$2,560 ~m area at 10~meter pixel resolution. 
Above-Ground Biomass (AGB) values were determined using open forest airborne LiDAR data and an extensive network of field plots as reference measurements by the Finnish Forest Centre. 
Patch coordinates and collection years--provided by the authors of Ref.~\cite{Nascetti} upon request--were used to retrieve satellite imagery to generate TESSERA and Google AlphaEarth Satellite Embeddings (AlphaEarth)~\cite{SatelliteEmbeddingV1}.

\paragraph{UNet model architecture }

To obtain AGB estimates, we trained a convolutional neural network based
on the UNet architecture~\cite{ronnebergerUNetConvolutionalNetworks2015}, 
using TESSERA or AlphaEarth embeddings ~\cite{SatelliteEmbeddingV1} as
input. 
The model received 256$\times$256$\times$128 and 256$\times$256$\times$64 input patches for TESSERA and AlphaEarth, respectively, 
where the dimension denotes the number of channels.

The network followed a standard UNet structure with two encoder
and decoder stages. 
The encoder comprised two convolutional blocks with 256 and 512 channels, 
each consisting of two 3×3 convolutions followed by batch normalization, 
ReLU activation, and 2×2 max pooling. 
A 1024-channel bottleneck block was used prior to upsampling. 
The decoder mirrored the encoder, using transposed convolutions for
upsampling and skip connections from the encoder, followed by double-convolution blocks. 
A final 1×1 convolution produced the single-channel regression output. 
The model had approximately 30 million trainable parameters. 
The loss function was defined as the root mean squared error (RMSE) 
averaged over individual patches.

Following the evaluation protocol in Ref.~\cite{Nascetti}, 
RMSE was computed on a test set of 2773 patches and reported 
as the average RMSE per patch. For limited label testing, a
subset of labels was randomly chosen from the remaining data and randomly divided into training and validation sets (80:20). The UNet model was trained on 
the embeddings for 80 epochs, with the model achieving the lowest loss on the validation set selected as the final model. \Cref{fig:agb} shows the average RMSE calculated over five random splits, with
bars indicating the minimum–maximum range when it exceeds the size
of the dot. Training was performed with Adam Optimizer
(learning rate = 1e-4, batch size = 4). 
Task-specific training on the entire dataset (80\% training / 20\% validation) using a single NVIDIA H200 GPU and 8 CPU cores took approximately 5 hours for TESSERA embeddings and 4 hours for the AlphaEarth embeddings.

\paragraph{Baselines}

In addition to AlphaEarth, we compared the performance of TESSERA to
two other baselines: the winner of the BioMassters competition \cite{Nascetti} and
the SpectralGPT foundation model \cite{SpectralGPT} combined with a UPerNet~\cite{xiao2018unified}. 

For SpectralGPT, we used the implementation provided in  Ref.~\cite{Pangaea}, 
with the BioMassters satellite data as input, a lightweight
spatial-temporal encoder L-TAE \cite{Sainte} for temporal aggregation, and a
UPerNet as the regression head. 
This setup was found in~\cite{Pangaea} to outperform other
tested foundation models on the BioMassters dataset. 
To ensure consistency in the limited label setting, we replaced the default
sampling strategy with the same random splits used to evaluate TESSERA
and normalized the inputs with their mean and standard deviation. 
Although we initially experimented with using per patch RMSE as
the loss function to align with our evaluation metric, we ultimately
reverted to the MSE loss~\cite{Pangaea} because it resulted in
slightly better performance. 
Obtaining the results for the full dataset using a single NVIDIA H200 GPU took approximately 18 hours.

For evaluating the supervised baseline, we downloaded the 
pre-trained weights of the BioMassters winning model and ran inference on 
the test set, applying a post-processing step to clip predictions to 
a non-negative range. 
Although we attempted to retrain the model from scratch using two H200 
GPUs and a different CUDA version than in the original setup, 
this configuration led to inferior performance compared to the original 
pre-trained model. 
According to the BioMassters codebase, the original model 
was trained on two A100 40GB GPUs over the course of eight days, 
highlighting the greater time efficiency of TESSERA, which required only five hours.

\subsection{Interactive Habitat Mapping Tool}

\begin{figure}[t]
    \centering
    \includegraphics[width=\linewidth]{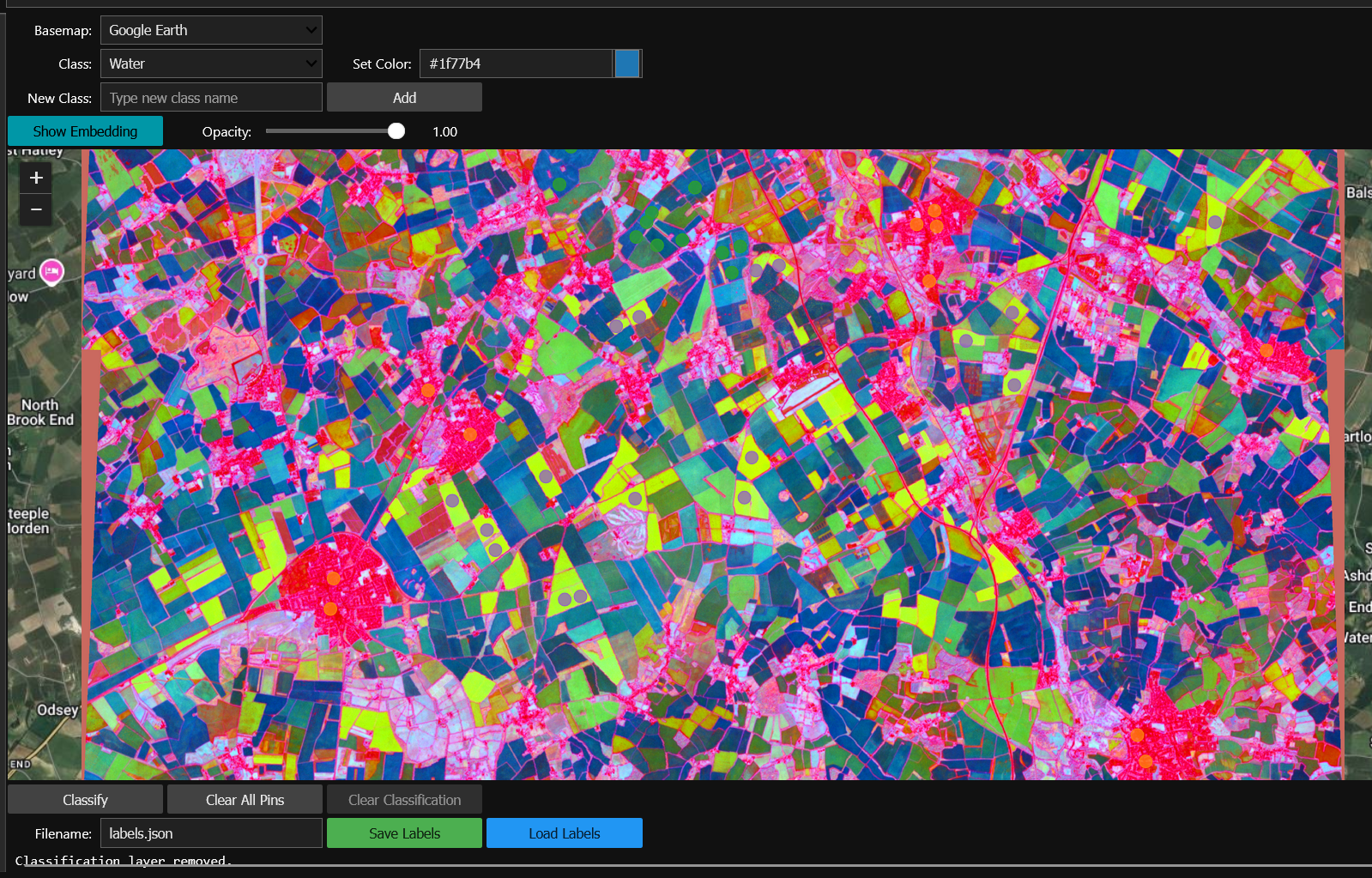}
    \caption{\textbf{We provide an interactive tool to allow non-specialist users to map TESSERA embeddings to user-defined classes.} We show a PCA-derived false colour map overlaid on an RGB aerial image of South Cambridge. }
    \label{fig:interactivetool}
\end{figure}

Here we demonstrate our interactive classification tool, a web-based utility designed for rapid human-in-the-loop habitat mapping using pre-computed model embeddings, as seen in ~\Cref{fig:interactivetool}. The tool allows a user to first define a region of interest, for which it fetches and mosaics the relevant high-dimensional embedding tiles. To aid visual interpretation, it performs Principal Component Analysis on the embeddings and displays the first three components as a false-colour RGB image overlay on a satellite basemap.

The core of the tool is the interactive labelling workflow. A user can define custom habitat classes and provide training data by clicking on the map to place labelled points. These points and their corresponding 128-dimensional embedding vectors are used to train a k-Nearest Neighbors (kNN) classifier. The trained model then predicts a class for every pixel in the region of interest, producing a map layer that overlays a coregistered aerial view. This process allows for iterative refinement, where a user can add more points to correct misclassifications and retrain the model for immediate feedback, demonstrating an accessible method for quickly creating habitat maps without requiring deep machine learning expertise. We also provide a batch processing tool for larger regions of interest. The tools and their open source code are freely available in the github repository.

\subsection{Pseudo Code}

This section provides the pseudocode for mixup regularization and global shuffling.

\makeatletter
\renewcommand{\ALG@beginalgorithmic}{\footnotesize}
\makeatother

\begin{algorithm}[t]
\footnotesize
\setlength{\lineskip}{1pt}
\caption{Mixup Consistency Regularization for Multimodal Barlow Twins}
\label{alg:mixup}
\begin{algorithmic}[1]

\State \texttt{def MixupBT(s2\_aug1, s2\_aug2, s1\_aug1, s1\_aug2, model, barlow\_lambda, lambd\_mix):}
\State \hspace{1em}\textcolor{blue}{\texttt{\# s2\_* and s1\_* are batches of augmented inputs}}
\State \hspace{1em}\textcolor{blue}{\texttt{\# model returns projection z and representation repr}}

\State \hspace{1em}\textcolor{blue}{\texttt{\# standard Barlow Twins branch}}
\State \hspace{1em}\texttt{z1, repr1 = model(s2\_aug1, s1\_aug1)}
\State \hspace{1em}\texttt{z2, repr2 = model(s2\_aug2, s1\_aug2)}
\State \hspace{1em}\texttt{loss\_bt, bar\_bt, off\_bt = BarlowTwins(z1, z2)}

\State \hspace{1em}\textcolor{blue}{\texttt{\# mixup branch}}
\State \hspace{1em}\texttt{B = s2\_aug1.shape[0]}
\State \hspace{1em}\texttt{idx = randperm(B)}
\State \hspace{1em}\texttt{alpha = Beta(beta\_alpha, beta\_beta).sample()}

\State \hspace{1em}\textcolor{blue}{\texttt{\# mix S2 and S1 views}}
\State \hspace{1em}\texttt{y\_m\_s2 = alpha * s2\_aug1 + (1 - alpha) * s2\_aug2[idx]}
\State \hspace{1em}\texttt{y\_m\_s1 = alpha * s1\_aug1 + (1 - alpha) * s1\_aug2[idx]}
\State \hspace{1em}\texttt{z\_m, repr\_m = model(y\_m\_s2, y\_m\_s1)}

\State \hspace{1em}\textcolor{blue}{\texttt{\# cross-correlation of mixed and original views}}
\State \hspace{1em}\texttt{cc\_m\_a   = Corr(z\_m,      z1)}
\State \hspace{1em}\texttt{cc\_m\_b   = Corr(z\_m,      z2)}
\State \hspace{1em}\texttt{cc\_11    = Corr(z1,        z1)}
\State \hspace{1em}\texttt{cc\_2idx1 = Corr(z2[idx],   z1)}
\State \hspace{1em}\texttt{cc\_12    = Corr(z1,        z2)}
\State \hspace{1em}\texttt{cc\_2idx2 = Corr(z2[idx],   z2)}

\State \hspace{1em}\textcolor{blue}{\texttt{\# mixup targets and consistency loss}}
\State \hspace{1em}\texttt{cc\_m\_a\_gt = alpha * cc\_11    + (1 - alpha) * cc\_2idx1}
\State \hspace{1em}\texttt{cc\_m\_b\_gt = alpha * cc\_12    + (1 - alpha) * cc\_2idx2}
\State \hspace{1em}\texttt{diff\_a     = (cc\_m\_a - cc\_m\_a\_gt).pow(2).sum()}
\State \hspace{1em}\texttt{diff\_b     = (cc\_m\_b - cc\_m\_b\_gt).pow(2).sum()}
\State \hspace{1em}\texttt{loss\_mix   = lambd\_mix * barlow\_lambda * (diff\_a + diff\_b)}

\State \hspace{1em}\textcolor{blue}{\texttt{\# final SSL loss with mixup consistency}}
\State \hspace{1em}\texttt{loss = loss\_bt + loss\_mix}
\State \hspace{1em}\texttt{return loss}

\end{algorithmic}
\end{algorithm}


\begin{algorithm}[t]
\footnotesize
\setlength{\lineskip}{1pt}
\caption{Global Shuffling and Multimodal Data Construction}
\label{alg:global_shuffling}
\begin{algorithmic}[1]

\State \texttt{def GlobalShuffling(tile\_roots, batch\_size, T):}
\State \hspace{1em}\textcolor{blue}{\texttt{\# tile\_roots: list of paths to MGRS tile directories}}
\State \hspace{1em}\textcolor{blue}{\texttt{\# batch\_size: number of tiles to process in memory simultaneously}}
\State \hspace{1em}\textcolor{blue}{\texttt{\# T: number of time steps for sparse sampling}}

\State \hspace{1em}\texttt{tiles = shuffle(tile\_roots)} \textcolor{blue}{\texttt{\# Randomize tile processing order}}
\State \hspace{1em}\texttt{num\_batches = len(tiles) // batch\_size}

\State \hspace{1em}\texttt{for b in range(num\_batches):}
\State \hspace{2.7em}\texttt{current\_tiles = tiles[b*batch\_size : (b+1)*batch\_size]}
\State \hspace{2.7em}\texttt{pixel\_buffer\_aug1 = []}
\State \hspace{2.7em}\texttt{pixel\_buffer\_aug2 = []}

\State \hspace{2.7em}\textcolor{blue}{\texttt{\# Parallel processing of tiles to extract valid pixels}}
\State \hspace{2.7em}\texttt{parfor tile in current\_tiles:}
\State \hspace{4.4em}\texttt{s2, s1, masks = load\_data(tile)}
\State \hspace{4.4em}\texttt{valid\_locs = filter\_valid\_pixels(masks, s1, s2)}
\State \hspace{4.4em}\texttt{if len(valid\_locs) == 0: continue}

\State \hspace{4.4em}\textcolor{blue}{\texttt{\# Sparse random temporal sampling per pixel}}
\State \hspace{4.4em}\texttt{for (y, x) in valid\_locs:}
\State \hspace{6.1em}\texttt{idx = get\_valid\_time\_indices(masks, y, x)}
\State \hspace{6.1em}\texttt{v1 = Sample(s2, s1, idx, T) \textcolor{blue}{\# View 1}}
\State \hspace{6.1em}\texttt{v2 = Sample(s2, s1, idx, T) \textcolor{blue}{\# View 2}}
\State \hspace{6.1em}\texttt{pixel\_buffer\_aug1.append(v1)}
\State \hspace{6.1em}\texttt{pixel\_buffer\_aug2.append(v2)}

\State \hspace{2.7em}\textcolor{blue}{\texttt{\# Aggregate pixels from spatially disjoint tiles}}
\State \hspace{2.7em}\texttt{stack\_aug1 = Stack(pixel\_buffer\_aug1) \textcolor{blue}{\# Shape: (N\_total, T, C)}}
\State \hspace{2.7em}\texttt{stack\_aug2 = Stack(pixel\_buffer\_aug2)}

\State \hspace{2.7em}\textcolor{blue}{\texttt{\# Global Shuffling: Decorrelate spatial neighborhoods}}
\State \hspace{2.7em}\texttt{N\_total = stack\_aug1.shape[0]}
\State \hspace{2.7em}\texttt{perm\_idx = randperm(N\_total)}
\State \hspace{2.7em}\texttt{shuffled\_aug1 = stack\_aug1[perm\_idx]}
\State \hspace{2.7em}\texttt{shuffled\_aug2 = stack\_aug2[perm\_idx]}

\State \hspace{2.7em}\textcolor{blue}{\texttt{\# Save as chunks for pre-training dataloader}}
\State \hspace{2.7em}\texttt{save\_chunks(shuffled\_aug1, shuffled\_aug2, out\_dir)}

\end{algorithmic}
\end{algorithm}

\newpage
\onecolumn
\begin{longtable}{lccccccccccc} 
\caption{\textbf{TESSERA provides unprecedented ease of use, scale, and accuracy.} Unlike other remote sensing foundation models, TESSERA provides analysis-ready and gap-free outputs (R), is open (O), and provides global (G) annual (A) coverage at 10~m resolution (10~m). It uses spectral-temporal features (ST) at pixel level (P), is invariant to clouds (CI), provides a compressed representation (CR), is oriented to land features (LO), and includes both multispectral and radar information (MS). 
    Codes: \y = Yes, \n = No, ? = Unknown, \p = Potentially true, .. = N/A.}\\
	\label{tab:sup_comparison} 
\textbf{Foundation Model} &  R & O & G & A & 10~m & ST & P & CI & CR & LO & MS \\
\hline
\endfirsthead
\hline
\textbf{Foundation Model }&  R & O & G & A & 10~m & ST & P & CI & CR & LO & MS \\
\hline
\endhead
\hline
\multicolumn{11}{r}{\textit{Continued on next page}} \\
\hline
\endfoot
\hline
\endlastfoot
TESSERA (2025)                           & \y & \y & \y & \y & \y & \y & \y & \y & \y & \y& \y\\ 
\hline

GeoKR (2021) \cite{GeoKR}               & \n & ? & \n & \n & \y & \n & \y & \n & \n & \y& \n\\ 
CVPRW (2021) \cite{Stojnic_2021_CVPR}               & \n & .. & \n & \n & \y & \n & \n& \n & .. & \y & \n\\
GASSL (2021) \cite{Ayush2021}               & \n & .. & \n & \n & \y & \n & \y & \n & .. & \y& \n \\
SeCo (2021) \cite{Manas2021}                & \n & \n & \n & \n & \y & \n & \n& \n & ? & \y& \n\\
MOSAIKS (2021) \cite{rolf2021generalizable}  & \n & \y & \y & \n & \n & \n & \n & \n & \y & \y& ? \\ 
            DINO-MM (2022) \cite{wang2022selfsupervisedvisiontransformersjoint}             & \n & .. & \n & \n & \y & \n & \n & \n & .. & \y& \y \\
            SatMAE (2022) \cite{Cong2022}              & \n & .. & \n & \n & \y & \y & \n & \p & .. & \p & \n \\
            RS-BYOL (2022) \cite{9880533}             & \n & .. & .. & \n & \y & \n & \y & \y & .. & \n & \y \\
            GeCo (2022)  \cite{ 9869651}               & \n & .. & .. & \n & .. & \n & \n & \n & .. & \p & \n \\
            RingMo (2022) \cite{Sun2023}              & \n & .. & .. & \n & \y & \n & \n & \n & ? & \y & \n \\
            RVSA (2022) \cite{9956816}                & \n & .. & .. & \n & .. & \n & \n & \n & .. & \y & \n \\
            RSP (2022) \cite{Wang2023}                 & \n & .. & .. & \n & \p & \n & \n & \p & \y & \y & \n \\
            MATTER (2022) \cite{Akiva2022}              & \n & .. & .. & \n & \y & \y & \n & \n & \y & \y & \n \\
            CSPT (2022) \cite{rs14225675}                & \n & .. & .. & \n & ? & \n & \n & \n & \y & \y & \n \\
            CVPRW (2022) \cite{Scheibenreif_2022_CVPR}               & \n & .. & .. & \n & \y & \n & \n & \n & \y & \y & \y \\
            BFM (2023) \cite{Cha_2024}                 & \n & .. & .. & \n & ? & \n & \n & \n & ? & \y & \n \\
            TOV (2023) \cite{ 10110958}                 & ? & .. & \n & \n & \y & \n & ? & \n & ? & \y & \n \\
            CMID (2023) \cite{10105625}                & \n & .. & \n & \n & \y & \n & \n & \n & ? & \y & \n \\
            RingMo-Sense (2023) \cite{10254320}        & \n & .. & \n & \n & \y & \y & \n & \n & ? & \y & \y \\
            lal-SimCLR (2023) \cite{ Prexl_2023_CVPR}          & \n & .. & \n & \n & \y & \n & \n & \n & \y & \y & \y \\
            CACo (2023) \cite{Mall_2023_CVPR}                & \n & .. & \n & \n & \y & \y & \n & \n & \y & \y & \n \\
            SatLas (2023) \cite{Bastani2023}              & \n & .. & \y & \n & \y & \y & \n & \n & \y & \y & \n \\
            GFM (2023) \cite{Mendieta2023}                 & \n & .. & \n & \n & \y & \n & \n & \n & \n & \y & \n \\
            Scale-MAE (2023) \cite{Reed2023}           & \n & .. & \n & \n & \y & \n & \n & \n & \n & \y & \n \\
            DINO-MC (2023) \cite{ wanyan2024extendinggloballocalviewalignment}             & \n & .. & \n & \n & \y & \n & \n & \n & \n & \y & \n \\
            CROMA (2023) \cite{fuller2023croma}               & \n & .. & \n & \n & \y & \n & \n & \n & \n & \y & \y \\
            Cross-Scale MAE (2023) \cite{Tang2023}     & \n & .. & \n & \n & \y & \n & \n & \n & \n & \y & \n \\
            DeCUR (2023) \cite{wang2024decouplingcommonuniquerepresentations}               & \n & .. & \n & \n & \y & \n & \n & \y & \n & \y & \y \\
            Presto (2023) \cite{tseng_lightweight_2024}              & \n & .. & \y & \y & \y & \y & \y & \y & \y & \y & \y \\
            CtxMIM (2023) \cite{zhang2024ctxmimcontextenhancedmaskedimage}              & \n & .. & \n & \n & \y & \n & \n & \n & \n & \y & \n \\
            FG-MAE (2023) \cite{ wang2023featureguidedmaskedautoencoder}              & \n & .. & \n & \n & \y & \n & \n & \n & \n & \y & \n \\
            Prithvi (2023) \cite{schmude2024prithviwxcfoundationmodel}             & \n & .. & \n & \n & \n & \y & \n & \y & \y & \y & \n \\
            RingMo-Lite (2023) \cite{ wang2023ringmoliteremotesensingmultitask}         & \n & .. & \n & \n & \y & \n & \n & \n & \n & \y & \n \\
            IGARSS (2023) \cite{ 10282433}              & \n & .. & \n & \n & \y & \n & \n & \n & \n & \y & \n \\
            EarthPTi (2023) \cite{smith2024earthpttimeseriesfoundation}            & \n & .. & \n & \n & \y & \y & \y & \y & \y & \y & \n \\
            USat (2023) \cite{irvin2023usatunifiedselfsupervisedencoder}                & \n & .. & \n & \n & \y & \n & \n & \n & \n & \y & \n \\
            AIEarth (2023) \cite{xu2023analyticalinsightearthcloudplatform}             & \p & ?  & \y & \n & \y & \n & \n & \n & \n & \y & \n \\
            Clay (2023) \cite{clay-foundation-model-2023}                & \n & .. & \y & \n & \y & \n & \n & \n & \n & \y & \n \\
            Hydro (2023) \cite{ Corley:2024}               & \n & .. & \n & \n & \y & \n & \n & \n & \n & \n & \n \\
            U-Barn (2023) \cite{10414422}              & \n & .. & \n & \y & \y & \y & \n & \y & \n & \y & \n \\
            GeRSP (2023) \cite{huang2024generic}               & \n & .. & \n & \n & \y & \n & \n & \n & \n & \y & \n \\
            SwiMDiff (2023) \cite{ tian2024swimdiffscenewidematchingcontrastive}            & \n & .. & \n & \n & \y & \n & \n & \n & \n & \y & \n \\
            OFA-Net (2023) \cite{xiong2024allunifiedfoundationmodels}             & \n & .. & \y & \n & \y & \n & \n & \n & \n & \y & \y \\
            SML-FR (2023) \cite{10378718}              & \p & \p  & \y & \n & \y & \n & \n & \n & \n & \n & \n \\
            Spectral-GP (2024) \cite{SpectralGPT}         & \n & \n & \y & \n & \y & \n & \n & \n & \n & \y & \n \\
            S2MAE (2024) \cite{li2024s2mae}               & \n & \n & \y & \n & \y & \n & \n & \n & \n & \y & \n \\
            SatMAE++ (2024) \cite{Noman2024}            & \n & \n & \y & \n & \y & \n & \n & \n & \n & \y & \n \\
            msGFM (2024) \cite{Han2024}               & \n & \n & \n & \n & \y & \n & \n & \n & \n & \y & \n \\
            SkySense (2024) \cite{Guo2024}            & \n & \n & \n & \n & \y & \y & \n & \n & \n & \y & \y \\
            MTP (2024) \cite{ wang2024mtpadvancingremotesensing}                 & \n & .. & \y & \n & \y & \n & \y & \n & \n & \y & \n \\
            DOFA (2024) \cite{ xiong2024neuralplasticityinspiredmultimodalfoundation}                & \n & .. & \y & \n & \y & \n & \n & \n & \y & \y & \y \\
            MMEarth (2024) \cite{Nedungadi2024}             & \n & .. & \y & \n & \y & \n & \n & \n & \y & \y & \y \\
            LeMeViT (2024) \cite{jiang2024lemevitefficientvisiontransformer}             & \n & .. & \n & \n & ? & \n & \n & \n & \y & \n & \n \\
            SoftCon (2024) \cite{ 10726860}             & \n & .. & \y & \n & \y & \n & \y & \n & \n & \y & \n \\
            RS-DFM (2024) \cite{ wang2024rsdfmremotesensingdistributed}              & \n & .. & \n & \n & \y & \n & \y & \n & \n & \y & \n \\
            A2MAE (2024) \cite{zhang2024a2maespatialtemporalspectralunifiedremote}               & \n & .. & \y & \n & \y & \n & \n & \n & \y & \y & \n \\
            HyperSIGM (2024) \cite{wang2025hypersigmahyperspectralintelligencecomprehension}           & \n & .. & \y & \n & \y & \y & \y & \n & \y & \y & \n \\
            SelectiveMAE (2024) \cite{wang2025harnessingmassivesatelliteimagery}        & \n & .. & \y & \n & \y & \n & \n & \n & \y & \y & \n \\
            OmniSat (2024) \cite{astruc2024omnisat}             & \n & .. & \n & \y & \y & \y & \n & ? & \y & \y & \y \\
            MM-VSE (2024) \cite{ravirathinam2024causally}              & \n & .. & \y & \y & \y & \y & \n & ? & \y & \y & \y \\
            MA3E (2024) \cite{Li2025}                & \n & .. & \n & \n & \y & \n & \n & \n & \y & \y & \n \\
            SpectralEarth (2024) \cite{braham2025spectralearth}       & \n & .. & \y & \y & \n & ? & \y & \n & \y & \y & \n \\
            SenPa-MAE (2024) \cite{prexl2024senpamaesensorparameteraware}           & \n & .. & \y & \y & \y & \y & \y & \n & \y & \y & \n \\
            RingMo-Ae (2024) \cite{diao2025ringmoaerialaerialremotesensing}           & \n & .. & \n & \n & \y & \n & \y & \n & \y & \y & \n \\      
            SAR-JEPA (2024) \cite{LI2024326}            & \n & .. & \n & \n & \y & \n & \n & \y & \n &\n & \n \\
            PIS (2024) \cite{10697182}            & \n & ? & \y & \y & \y & .. & .. & .. & .. & .. & .. \\
            OReole-FM (2024) \cite{Dias_2024}            & \n & .. & \y & \n & \y & \n & \y & \n & \y & \y & \n \\
            PIEVIT (2024) \cite{lu2025pattern}            & \n & .. & \y & \n & \y & \n & \y & \n & \y & \y & \n \\
            SatVisionTOA (2024) \cite{spradlin2024satvisiontoageospatialfoundationmodel}      & \n & .. & \y & \y & \n & \y & \y & \n & \y & \y & \n \\
            RS-vHeat (2024) \cite{hu2025rsvheatheatconductionguided}            & \n & .. & ? & ? & ? & ? & \y & ? & \y & ? & \y \\
            Prithvi-EO-2.0 (2024) \cite{szwarcman2025prithvieo20versatilemultitemporalfoundation}   & \n & .. & \y & \y & \n & \y & \n & \y & \y & \y & \n \\
            AnySat (2024) \cite{astruc2025anysatearthobservationmodel}            & \n & .. & \y & \n & \y & \n & \y & \n & \y & \y & \y \\
            WildSAT (2024) \cite{daroya2024wildsatlearningsatelliteimage}            & \n & \n & \n & \n & \n & \n & \n & \n & \y & \y & \n \\
            SeaMo (2024) \cite{li2025seamoseasonawaremultimodalfoundation}            & \n & \n & \n & \n & \y & \n & \n & \n & \y & \y & \n \\
            FoMo (2025) \cite{bountos2025fomomultimodalmultiscalemultitask}            & \y & \y & \y & \n & \y & \n & \y & \n & \y & \y & \y \\
            SatMamba (2025) \cite{duc2025satmambadevelopmentfoundationmodels}            & \y & \y & \n & \n & \n & \n & \n & \n & \n & \y & \n \\
            Galileo (2025) \cite{tseng2025galileolearningglobal}            & \p & \p & \y & ? & \y & \n & ? & ? & \y & \y & \y \\
            SatDiFuser (2025)\cite{jia2025generativegeospatialdiffusionmodels}            & \n & \n & \n & \n & \y & \n & \n & \n & \n & \y & \n \\
            RoMA (2025) \cite{wang2025romascalingmambabasedfoundation}            & \n & \n & \n & \n & \y & \n & \n & \n & \n & \y & \n \\
            Panopticon (2025) \cite{waldmann2025panopticonadvancinganysensorfoundation}       & \n & \y & \n & \n & \y & \y & \n & \n & \y & \y & \y \\
            FedSense (2025) \cite{tan2025towards}            & \n & \n & \n & \n & \n & \n & \n & \n & \n & \y & \n \\
            Copernicus-FM (2025) \cite{wang2025unifiedcopernicusfoundationmodel}            & \n & .. & \y & \n & \y & \y & \n & \n & \y & \y & \y \\
            HyperFree (2025) \cite{li2025hyperfree}            & \n & .. & \n & \n & \y & \n & \n & \n & \y & \y & \n \\
            DynamicVis (2025) \cite{chen2025dynamicvisefficientgeneralvisual}            & \n & .. & \n & \n & \y & \n & \n & \n & \n & \y & \y \\
            FlexiMo (2025) \cite{li2025fleximoflexibleremotesensing}            & \n & .. & \n & \n & \y & \n & \n & \n & \n & \y & \y \\
            Google Satellite (2025)\cite{SatelliteEmbeddingV1} & \y & \n & \y & \y & \y & \y & \n & \y & \y  & \y & \y \\
            SkySense++ (2025) \cite{wu2025semantic}            & \n & \y & \y & \p & \p & \y & \n & \p & \y & \y & \y \\
            RingMoE (2025) \cite{bi2025ringmoemixtureofmodalityexpertsmultimodalfoundation}       & \n & .. & \n & \n & \y & \y & \n & \n & \y & \y & \y \\       
		\hline
\end{longtable}

\end{document}